\documentclass[Afour,sageh,times]{sagej}

\usepackage{moreverb,url}

\usepackage[colorlinks,bookmarksopen,bookmarksnumbered,citecolor=black,urlcolor=black]{hyperref}
\usepackage{graphicx}
\usepackage{natbib}
\usepackage{caption}
\usepackage{subcaption}
\usepackage{amsmath}
\DeclareMathOperator*{\argmax}{arg\,max}

\newcommand\figref {Figure~\ref}
\newcommand{\eref}{Equation~\eqref}
\usepackage{wrapfig}
\usepackage{amssymb}
\usepackage{booktabs}
\usepackage{multirow}
\usepackage{graphicx}
\usepackage{paralist}
\usepackage[font=scriptsize,labelfont=sl,labelsep=period]{caption}
\usepackage{subfiles}
\usepackage{array}

\newcommand{\Rot}{\mathrm{Rot}}
\newcommand{\SOtwo}{\mathrm{SO}(2)}
\newcommand{\reg}{\mathrm{reg}}
\newcommand{\quot}{\mathrm{quot}}
\newcommand{\irrep}{\mathrm{irrep}}
\newtheorem{proposition}{Proposition}

\newcommand{\SE}{\mathrm{SE}}
\newcommand{\SO}{\mathrm{SO}}
\newcommand{\pick}{\mathrm{pick}}
\newcommand{\place}{\mathrm{place}}

\usepackage{soul}
\usepackage{xcolor}

\newcommand\BibTeX{{\rmfamily B\kern-.05em \textsc{i\kern-.025em b}\kern-.08em
T\kern-.1667em\lower.7ex\hbox{E}\kern-.125emX}}

\def\volumeyear{2016}

\newtheorem{lemma}{Lemma}

\usepackage{enumitem}
\usepackage{gensymb}
\begin{document}

\runninghead{Huang, Wang, Tangri, Walters and Platt}

\title{Leveraging Symmetries in Pick and Place}

\author{Huang, Haojie and Wang, Dian and Tangri, Arsh and Walters, Robin\affilnum{$\ast$} and Platt, Robert\affilnum{$\ast$}}
\affiliation{
\affilnum{$\ast$} Equal Advising \\
Khoury College of Computer Science, Northeastern University, Boston, USA \\
This work was supported in part by NSF 1724257, NSF 1724191, NSF1763878, NSF 1750649, NSF 2107256, NSF 2134178, NASA 80NSSC19K1474, the Harold Alfond Foundation, and the Roux Institute.}
\corrauth{Haojie Huang, Northeastern University, Boston, MA 02115}
\email{huang.haoj@northeastern.edu}

\begin{abstract}
Robotic pick and place tasks are symmetric under translations and rotations of both the object to be picked and the desired place pose. For example, if the pick object is rotated or translated, then the optimal pick action should also rotate or translate. The same is true for the place pose; if the desired place pose changes, then the place action should also transform accordingly. 
A recently proposed pick and place framework known as Transporter Net~\citet*{zeng2021transporter} captures some of these symmetries, but not all. 
This paper analytically studies the symmetries present in planar robotic pick and place and proposes a method of incorporating equivariant neural models into Transporter Net in a way that captures all symmetries. The new model, which we call \emph{Equivariant Transporter Net}, is equivariant to both pick and place symmetries and can immediately generalize pick and place knowledge to different pick and place poses. We evaluate the new model empirically and show that it is much more sample efficient than the non-symmetric version, resulting in a system that can imitate demonstrated pick and place behavior using very few human demonstrations on a variety of imitation learning tasks.

\end{abstract}


\maketitle
\section{Introduction}
\begin{figure*}
     \centering
     \begin{subfigure}[b]{0.47\textwidth}
         \centering
         \includegraphics[width=0.7\textwidth]{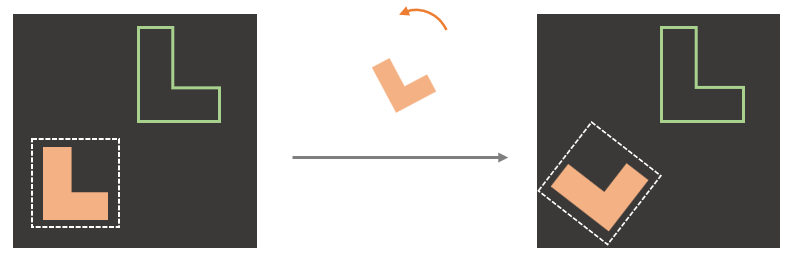}
         \caption{If Transporter Network~\cite{zeng2021transporter} learns to place an object when it is presented in one orientation, the model is immediately able to generalize to new object orientations.}
         \label{fig:transporter_symmetry}
     \end{subfigure}
     \hspace{0.5cm}
     \begin{subfigure}[b]{0.47\textwidth}
         \centering
         \includegraphics[width=0.7\textwidth]{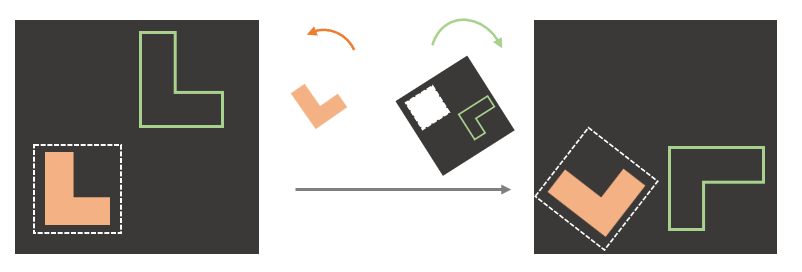}
         \caption{Our proposed Equivariant Transporter Network is able to generalize over both pick and place orientation. We view this as ${\SOtwo} \times {\SOtwo}$-place symmetry of the model.}
         \label{fig:place_symmetry}
     \end{subfigure}
     \caption{ Visual explanation of ${\SOtwo}$-equivariace (left figure) v.s. ${\SOtwo} \times {\SOtwo}$-equivariance of the place model (right figure)}
     \label{fig:cn_vs_cnxcn}
\end{figure*}
Pick and place is an important paradigm in robotic manipulation where a complex manipulation problem can be decomposed into a sequence of grasp (pick) and place operations. Recently, multiple learning approaches have been proposed to solve this problem, including~\citet*{zeng2021transporter, wang2021policy}. These methods focus on a simple version of the planar pick and place problem where the method looks at the scene and outputs a single pick and a single place pose. This problem has an important structure in the form of symmetries in $\SE(2)$ that can be expressed with respect to the pick and place pose. The pick symmetry is easiest to see. If the object to be grasped is rotated (in the plane), then the optimal grasp pose clearly must also rotate. A similar symmetry exists in place pose. If an object is to be placed into an environment in a particular way, then if the environment rotates, the desired place pose must also rotate. {Leveraging symmetries of the task could result in significant gains in sample efficiency~\cite{zhu2022grasp,jia2023seil}. Why is sample efficiency important in robot learning? Although robotic simulators could provide a huge amount of data that could be used to train a policy, there is an inevitable sim-to-real gap in applying the learned policy directly to real robots. On the other side, real-world robot data is expensive to collect, and sample efficiency is crucial to learning a policy with a limited number of human demonstrations.}

If we are to design a robotic learning system for pick and place, it should ideally encode the {symmetries} described above. This is a structure that exists in the problem and there is a possibility to simplify learning by encoding this structure into our learned solutions. The question is how to accomplish this. This paper examines the symmetries that exist in the pick and place problem by identifying invariant and equivariant equations that we would expect to be preserved. Then, we consider existing pick and place models and find that those architectures only express some but not all problem symmetries. Finally, we propose a novel pick and place model that we call \emph{Equivariant Transporter Net} that encodes all symmetries and shows that it outperforms models that do not preserve the relevant symmetries.

\subsection{Symmetries in Transporter Net}

This paper builds on top of the \emph{Transporter Net} model~\citet*{zeng2021transporter}. Transporter Net is a sample-efficient model for learning planar pick and place behaviors through imitation learning. Compared to many other approaches~\citet*{Qureshi2021NeRP,curtis2022long}, it does not need to be pre-trained on the involved objects -- it only needs to be trained on the given demonstrations. Transporter Net achieves sample efficiency in this setting by encoding {the symmetry of the picked object} into the model. Once the model learns to pick and place an object presented in one orientation, that knowledge immediately generalizes to a finite set of other pick poses. This is illustrated in Figure~\ref{fig:transporter_symmetry}. The left side of Figure~\ref{fig:transporter_symmetry} shows a pick-place problem where the robot must pick the orange object and place it inside the green outline. Because the model encodes {the symmetry of the picked object}, the ability to solve the place task on the left side of Figure~\ref{fig:transporter_symmetry} immediately implies an ability to solve the place task on the right side of Figure~\ref{fig:transporter_symmetry} where the object to be picked has been rotated. We will refer to this as a ${SO(2)}$-place symmetry. 
{{Since \emph{Transporter Net} used a set of discrete rotations, it actually achieves $C_n$-place symmetry where $C_n$ is the finite cyclic subgroup of $\SO(2)$ that contains a set of $n$ rotations.}}

\subsection{Equivariant Transporter Net}

This paper analyzes the symmetries present in the pick and place problem and expands Transporter Net in the following ways. First, we constrain the pick model to be equivariant (an expression of symmetry) with respect to the ${\SOtwo}$ group by incorporating equivariant convolutional layers into the pick model. This is, if there is a rotation on the object to be picked, the pick pose will also rotate. We refer to this as a ${\SOtwo}$-pick symmetry. The second way we extend Transporter Net is by making it equivariant with respect to changes in place orientation. That is, if the place model learns how to place an object in one orientation, that knowledge generalizes immediately to different place orientations. Our resulting placing model is equivariant both to changes in pick and place orientation, and can be viewed as a direct product of two groups, ${\SOtwo} \times {\SOtwo}$ as illustrated in Figure~\ref{fig:place_symmetry}. This expanded symmetry improves the sample efficiency of our model by enabling it to generalize over a larger set of problems. Finally, we also propose a goal-conditioned version of Equivariant Transporter Net where the desired place pose is provided to the system in the form of an image as shown in ~\figref{fig:goal-conditioned_task}.

\subsection{Contributions}

Our specific contributions are as follows. 1) We {systematically} analyze the symmetries present in the planar pick and place problem. 2) We propose Equivariant Transporter Net, a novel version of Transporter Net that has $C_n$-equivariant {pick symmetry} and $C_n \times C_n$-equivariant {place symmetry}.\footnote[1]{{Our implementation uses the discrete $C_n$ group instead of the continuous $\SOtwo$ group in order to compare with the baseline \emph{Transporter Net}~\cite{zeng2021transporter}. The $\SOtwo$ version of our model could be easily achieved with the irreducible representations based on our implementation.}} 3) We propose a {variation} of Equivariant Transporter Net that can be used with standard grippers rather than just suction cups. 4) We propose a goal-conditioned version of Equivariant Transporter Net. 5) We evaluate the approach both in simulation tasks and on physical robot versions of three of the gripper tasks. Our results indicate that our approach is more sample efficient than the baselines and therefore learns better policies from a small number of demonstrations. Video and code are available at \url{https://haojhuang.github.io/etp_page/}.

{This paper extends the recent work~\citet*{Huang-RSS-22} in the following ways. 
First, we cover the concepts, algorithms, and results in a more comprehensive way. Second, we generalize our proofs of equivariance from $C_n$ to any subgroup of $\SOtwo$. We also analyze the extension to $\SO(3)$ mathematically and provide intuition. 
Third, we propose a goal-conditioned extension of the work and show that the new method outperforms on the benchmark of goal-conditioned tasks. 
Finally, we add an ablation study that characterizes the model for differently sized cyclic groups, $C_n$.}

\subsection{Comparison to related works}

\textbf{Pick and Place.} Pick and place is an important topic in manipulation. Many fundamental skills like packing, kitting, and stacking require inferring both the pick and the place action. Traditional assembly methods in factories use customized workstations so that fixed pick and place actions can be manually predefined. Recently, considerable research has focused on vision-based manipulation. Some work~\citet*{narayanan2016discriminatively,chen2019grip,gualtieri2021robotic} assumes that object mesh models are available in order to run ICP~\citet*{besl1992method} and align the object model with segmented observations or completions~\citet*{yuan2018pcn,huang2021gascn}. Other work learns a category-level pose estimator~\citet*{yoon2003real,deng2020self} or key-point detector~\citet*{nagabandi2020deep,liu2020keypose, manuelli2019kpam} from training on a large dataset.{Recently, \citet*{Wen-RSS-22} realizes a close-loop intra-category policy by mimicking the extracted pose trajectory from a few video demonstrations.} However, these methods often require expensive object-specific labels {or pre-training}, making them difficult to use widely. Recent advances in deep learning have provided other ways to rearrange objects from perceptual data. 
\citet*{Qureshi2021NeRP} represent the scene as a graph over segmented objects to do goal-conditioned planning; 
\citet*{curtis2022long} propose a general system consisting of a perception module, grasp module, and robot control module to solve multi-step manipulation tasks. These approaches often require {prior knowledge like good segmentation module and human-level hierarchy}. End-to-end models~\citet*{zakka2020form2fit,khansari2020action,devin2020self,berscheid2020self} that directly map input observations to actions can learn quickly and generalize well. \citet*{shridhar2022cliport} learn one multi-task policy with language-conditioned imitation-learning.
{\citet*{shridhar2022peract} directly extend this idea to 3D keyframe multi-task policy learning with Perceiver IO transformer~\citet*{jaegle2021perceiver}.}
\citet*{wu2020learning} achieve fast learning speed on deformable object manipulation tasks with reinforcement learning. 
However, most methods need to be trained on large datasets. For example, 
\citet*{khansari2020action} collects a dataset with 7.2 million samples. 
\citet*{devin2020self} collects $40K$ grasps and places per task. 
\citet*{zakka2020form2fit} collects 500 disassembly sequences for each kit. The focus of this paper is on improving the sample efficiency of this class of methods {on various manipulation tasks.}

\vspace{0.3cm}
\textbf{Equivariance Learning in Manipulation.}
Fully Convolutional Networks (FCN) are translationally equivariant and have been shown to improve learning efficiency in many manipulation tasks~\cite*{zeng2018robotic,Morrison-RSS-18}. The idea of encoding SE(2) symmetries in the structure of neural networks is first introduced in G-Convolution~\citet*{cohen2016group}. The extension work proposes an alternative architecture, Steerable CNN~\citet*{cohensteerable}. 
\citet*{weiler2019general} propose a general framework for implementing E(2)-Steerable CNNs. { \citet*{weiler20183d} first investigated the SE(3) steerable convolution kernels for volumetric data with the trick of vectorizing. \citet*{cesa2021program} parameterizes filters with a band-limited basis to build E(3)-steerable kernels. \citet*{thomas2018tensor} and \citet*{fuchs2020se} extended the equivariance to graph neural networks.} 

In the context of robotics learning,
\citet*{zhu2022grasp} decouple rotation and translation symmetries to enable the robot to learn a planar grasp policy {online} within $1.5$ hours. {Compared with \citet*{zhu2022grasp} that formulated the planar grasping task as a bandit problem, our work focuses on pick-place tasks and learns from demonstrations.}
\citet*{wang2022equivariant} use SE(2) equivariance in Q learning to solve multi-step sequential manipulation {pick-place} tasks. {Compared with \citet*{wang2022equivariant}, our work leverages the larger $\SOtwo \times \SOtwo$ symmetry group for the pick-conditioned place policy and tackles rearrangement tasks through the imitation learning~\citet*{hussein2017imitation,hester2018deep,vecerik2017leveraging}.} Recently, various SE(3) equivariant acrhitectures~\cite*{thomas2018tensor,fuchs2020se,chen2021equivariant, deng2021vector} have been proposed and applied to solve manipulation problems.
\citet*{simeonov2022neural} use Vector Neurons~\cite*{deng2021vector} to get SE(3)-invariant object representations so that the model can manipulate objects in the same category with a few training demonstrations. 
\citet*{huang2022edge} leverages the SE(3) invariance of the grasping evaluation function to enable better grasping performance. 
\citet*{xue2022useek} use SE(3)-equivariant key points to infer the object's pose for pick and place.
{However, most SE(3)-equivariant pick-place methods~\citet*{simeonov2022neural,xue2022useek} require a segmentation model and a pre-trained point descriptor for each category, which limits their adaptations to various tasks.
Although our proposed pick-place symmetry is defined on SE(2) in this work, we will briefly analyze how to extend the idea to SE(3)-pick-place problems in Proposition~\ref{pro:steerable}.}

\section{Background on Symmetry Groups}
\label{sect:background}

\subsection{The Groups $\SOtwo$ and $C_n$}
{In this work}, we primarily {focus on} rotations expressed by the group $\SOtwo$ and its cyclic subgroup $C_n \subseteq \SOtwo$. $\SOtwo$ contains the continuous planar rotations $\{ \Rot_{\theta}: 0\leq \theta < 2\pi\}$. The discrete subgroup $C_n = \{ \Rot_{\theta}: \theta \in \{\frac{2\pi i}{n}| 0\leq i < n\} \}$ contains only rotations by angles which are multiples of $2\pi/n$. The special Euclidean group $\mathrm{SE}(2) = \SOtwo \times \mathbb{R}^2$ describes all translations and rotations of $\mathbb{R}^2$.

\subsection{Representation of a Group}

A $d$-dimensional \emph{representation} $\rho \colon G \to \mathrm{GL}_d$ of a group $G$ assigns to each element $g \in G$ an invertible $d\!\times\! d$-matrix $\rho(g)$.  Different representations of $\SOtwo$ or $C_n$ help to describe how different signals are transformed under rotations.  

\begin{enumerate}
    {
    \item \textbf{The trivial representation} $\rho_0 \colon \SOtwo \to \mathrm{GL}_1$ assigns $\rho_0(g) = 1$ for all $g \in G$, i.e. no transformation under rotation. 
    
    \item \textbf{The standard representation}
    \[\rho_1(\Rot_\theta) = \begin{pmatrix}
    \cos{\theta} & -\sin{\theta} \\
    \sin{\theta} & \cos{\theta}
    \end{pmatrix}
    \]
    represents each group element by its standard rotation matrix. Notice that $\rho_0$ and $\rho_1$ can be used to represent elements from either $\SO(2)$ or $C_n$.
    
    \item \textbf{The regular representation} $\rho_{\mathrm{reg}}$ of $C_n$ acts on a vector in $\mathbb{R}^{n}$ by cyclically permuting its coordinates $\rho_{\reg}(\Rot_{2\pi /n})(x_0,x_1,...,x_{n-2},x_{n-1})=(x_{n-1} ,x_0,x_1,...,x_{n-2})$. We can rotate by multiples of $2\pi/n$ by $\rho_{\reg}(\Rot_{2\pi i /n}) = \rho_{\reg}(\Rot_{2\pi /n})^i$.
    
    \item \textbf{The quotient representation} of $C_n$ for $k$ dividing $n$ is denoted $\rho_{\quot}^{C_n/C_k}$ and acts on $\mathbb{R}^{n/k}$ by permuting $|C_n|/|C_k|$ channels: $\rho_{\quot}^{C_n/C_k}(\Rot_{2\pi i /n})(\mathbf{x})_j = (\mathbf{x})_{j + i\  \mathrm{mod} (n/k)}$, which implies features that are invariant under the action of $C_k$. 

    \item \textbf{The irreducible representation} $\rho_{\mathrm{irrep}}^{i}$ could be considered as the basis function with the order/frequency of $i$, such that any representation $\rho$ of $G$ could be decomposed as a \emph{direct sum} of them: $\rho(g)=Q^{\top}(\bigoplus_{i}\rho_{\mathrm{irrep}}^{i})Q$, where $Q$ is an orthogonal matrix.
    }
\end{enumerate}
 
 For more details, we refer interesting readers to~\citet*{serre1977linear}, \citet*{weiler2019general}, {\citet*{lang2020wigner} and~\citet*{cesa2021program}.}

\subsection{Feature Map Transformations}

\begin{figure}
    \centering
    \includegraphics[width=0.18\textwidth]{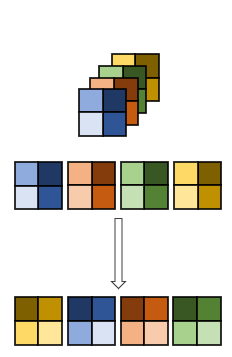}
    \caption{Illustration of the action of $T_g^{\reg}$ on a $2 \times 2$ image.}
    \label{fig:rotation_illustration}
\end{figure}
We formalize images and {2D} feature maps as feature vector fields, i.e. functions $f \colon \mathbb{R}^2 \rightarrow \mathbb{R}^c$, which assign a feature vector $f(\mathbf{x}) \in \mathbb{R}^c$ to each position $\mathbf{x} \in \mathbb{R}^2$. While in practice we discretize and truncate the domain of $f$ $\lbrace (i,j) : 1 \leq i \leq W, 1 \leq j \leq W \rbrace$, here we will consider it to be continuous for the purpose of analysis. The action of {an element} $g \in \SOtwo$ on $f$ is a combination of a rotation in the domain of $f$ via $\rho_1$ (this rotates the pixel positions) and a {transformation} in the channel space $\mathbb{R}^c$ {(aka. fiber space)} by {$\rho \in \{\rho_0,\rho_1,\rho_{\reg},\rho_{\irrep}\}$.} If $\rho = \rho_{\reg}$, then the channels cyclically permute according to the rotation. If $\rho = \rho_0$, the channels do not change. We denote this action (the action of $g$ on $f$ via $\rho$) by $T^{\rho}_g(f)$:
\begin{equation}
\label{eqn:rotation_illustration}
[T^{\rho}_g(f)](\mathbf{x}) = \rho(g) \cdot f( \rho_1(g)^{-1} \mathbf{x}).
\end{equation}
For example, the action of $T^{\rho_{\reg}}_g(f)$ is illustrated in Figure~\ref{fig:rotation_illustration} for a rotation of $g = \pi/2$ on a $2 \times 2$ image $f$ that uses $\rho_{\reg}$. The expression $\rho_1(g)^{-1} \mathbf{x}$ rotates the pixels via the standard representation. Multiplication by $\rho(g) = \rho_{\reg}(g)$ permutes the channels. For brevity, we will denote $T_g^{\reg} = T_g^{\rho_{\reg}}$ and $T_g^{0} = T_g^{\rho_{0}}$.

\subsection{Equivariant Mappings {and Steerable Kernels}}

A function $F$ is equivariant if it commutes with the action of the group,
\begin{equation}
\label{eqn:equivariance}
    T_g^{\mathrm{out}}(F(f)) = F(T_g^{\mathrm{in}}(f))
\end{equation}
where $T_g^{\mathrm{in}}$ transforms the input to $F$ by the group element $g$ while $T_g^{\mathrm{out}}$ transforms the output of $F$ by $g$. For example, if $f$ is an image, then  $\SO(2)$-equivariance of $F$ implies that it acts on $f$ in the same way regardless of the orientation in which $f$ is presented. That is, if $F$ takes an image $f$ rotated by $g$ (RHS of Equation~\ref{eqn:equivariance}), then it is possible to recover the same output by evaluating $F$ for the un-rotated image $f$ and rotating its output (LHS of Equation~\ref{eqn:equivariance}). {The most equivariant mappings between spaces of feature fields are \emph{convolutions with G-steerable kernels}~\cite{weiler20183d,jenner2021steerable}. Denote the input field type as $\rho_{\mathrm{in}}\colon G \rightarrow \mathbb{R}^{d_{\mathrm{in}}\times d_{\mathrm{in}}}$ and the output field type as $\rho_{\mathrm{out}}\colon G \rightarrow \mathbb{R}^{d_{\mathrm{out}}\times d_{\mathrm{out}}}$. The G-steerable kernels are convolution kernels $K\colon \mathbb{R}^{n} \rightarrow \mathbb{R}^{d_{\mathrm{out}}\times d_{\mathrm{in}}}$ satisfying the \emph{steerability constraint}, where $n$ is the dimensionality of the space
\begin{equation}
    K(g\cdot x) = \rho_{\mathrm{out}}(g)K(x)\rho_{\mathrm{in}}(g)^{-1}
    \label{equ:steerablility_constraint}
\end{equation}
}

\section{Problem Statement}

{This paper considers behavior cloning for planar pick and place problems. These problems are planar in the sense that the observation is a top-down image and the pick and place actions are motions to coordinates in the plane. Given a set of demonstrations that contains a sequence of one or more observation-action pairs $(o_t, a_t)$, the objective is to infer a policy $p(a_t|o_t)$ where the action $a_t = (a_{\pick}, a_{\place})$ describes both the pick and place components of action, and the observation $o_t$ describes the current state in terms of a top-down image of the workspace.}

 {Our model will encode this policy by factoring $p(a_{\pick} | o_t)$ and $p(a_{\place}|o_t,a_{\pick})$ and representing them as two separate neural networks. This policy be can used to solve tasks that are solvable in a single time step (i.e. a single pick and place action) as well as tasks that require multiple pick and place actions to solve. $a_{\pick}$ and $a_{\place}$ are parameterized in terms of $\SE(2)$ coordinates $(u,v,\theta)$, where $u,v$ denote the pixel coordinates of the gripper position and $\theta$ denotes gripper orientation. $\theta_{\pick}$ is defined with respect to the world frame and $\theta_{\place}$ is the delta action between the pick pose and place pose.}

\section{Transporter Network}

Before describing Equivariant Transporter Net, we analyze the original Transporter Net~\citet{zeng2021transporter} architecture from a different perspective.

\subsection{Description of Transporter Net}
\label{sect:transporter_desc}

\begin{figure}[t]
    \centering
    \includegraphics[width=0.5\textwidth]{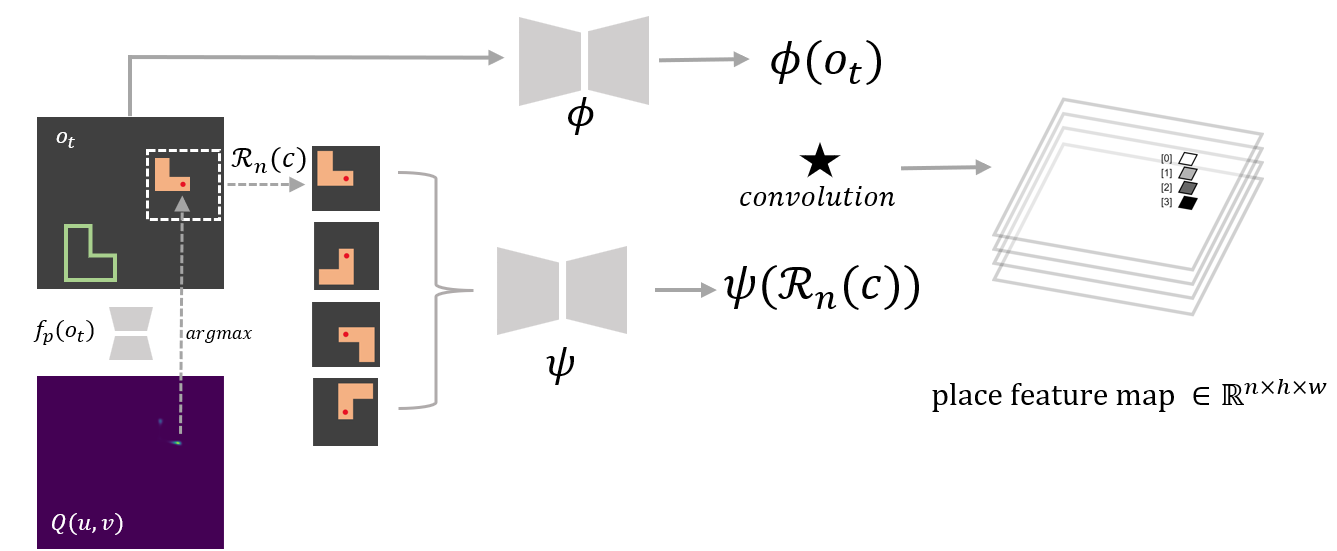}
    \caption[Transporter Network]{The Architecture of Transporter Net.}
    \label{fig:transporter_architecture}
\end{figure}

Transporter Network~\cite{zeng2021transporter} solves the planar pick and place problem using the architecture shown in Figure~\ref{fig:transporter_architecture}. The pick network $f_{\pick} \colon o_t \mapsto p(u,v)$ maps and image $o_t$ onto a probability distribution $p(u,v)$ over pick position $(u,v) \in \mathbb{R}^2$. The output pick position $a_{\pick}^*$ is calculated by maximizing $f_{\pick}(o_t)$ over $(u,v)$. (Since~\cite{zeng2021transporter} uses suction cups to pick, that work ignores pick orientation.) The place position and orientation is calculated as follows. First, an image patch $c$ centered on $a_{\pick}^*$ is cropped from $o_t$ to represent the pick action as well as the object. Then, the crop $c$ is rotated $n$ times to produce a stack of $n$ rotated crops. We denote this stack of crops as

\begin{equation}
    \mathcal{R}_n(c) = (T^0_{2\pi i/n}(c))_{i=0}^{n-1},
\end{equation}
where we refer to $\mathcal{R}_n$ as the ``lifting'' operator {of $C_n$}. Then, $\mathcal{R}_n(c)$ is encoded using a neural network $\psi$.
The original image, $o_t$, is encoded by a separate neural network $\phi$. The distribution over place location is evaluated by taking the cross correlation between $\psi$ and $\phi$,
\begin{equation}
\label{eqn:transporter1}
    f_{\place}(o_t,c) = \psi(\mathcal{R}_n(c)) \star \phi(o_t),
\end{equation}
where $\psi$ is applied independently to each of the rotated channels in $\mathcal{R}_n(c)$.  Place position and orientation is calculated by maximizing $f_{\place}$ over the pixel position (for position) and the orientation channel (for orientation).

\subsection{Analysis of Transporter Net}

The model architecture described above gives Transporter Network the following equivariance property.

\begin{proposition}
\label{prop:equivtransporter}
The Transporter Net place network $f_{\place}$ is $C_n$-equivariant. That is, given $g \in C_n$, object image crop $c$ and scene image $o_t$, 
\begin{equation}
    \label{eqn:transporter3}
    f_{\place}(o_t,T^0_g(c))  = \rho_{\reg}(-g) f_{\place}(o_t,c).
\end{equation}
\end{proposition}

Proposition~\ref{prop:equivtransporter} expresses the following intuition. A rotation of $g$ applied to the orientation of the object to be picked results in a $-g$ change in the placing angle, which is represented by a permutation along the channel axis of the placing feature maps. We denote the permutation in the channel space as $\rho_{\reg}(-g)$. This is a symmetry over the cyclic group $C_n \subseteq \SO(2)$ which is encoded directly into the model. It enables it to immediately generalize over different orientations of the object to be picked and thereby improves sample efficiency.

To prove Proposition 1,  We start with some common lemmas. In order to understand continuous rotations of image data, it is helpful to consider a $k$-channel image as a mapping $f\colon \mathbb{R}^2 \to \mathbb{R}^k$ where the input $\mathbb{R}^2$ defines the pixel space.  We consider images centered at $(0,0)$ and for non-integer values $(x,y)$ we consider $f(x,y)$ to be the interpolated pixel value.  Similarly, let $K: \mathbb{R}^2 \to \mathbb{R}^{l\times k}$ be a convolutional kernel where $k$ is the number of the input channels and $l$ is the number of the output channels.  Although the input space is $\mathbb{R}^2$, we assume the kernel is $r\!\times\!r$ pixels and $K(x,y)$ is zero outside this set. The convolution can then be expressed by
  $(K\star f)(\vec{v})= \sum_{\vec{w}\in \mathbb{Z}^2}f(\vec{v}+\vec{w})K(\vec{w })$, where $\vec{v}=(i,j)\in\mathbb{R}^2$.

 {Without loss of generality, assume that $f \colon \mathbb{R}^2 \to \mathbb{R}$ and define $\tilde f \colon \mathbb{R}^2 \to \mathbb{R}^{n}$ to be the $n$-fold duplication of $f$ such that $\tilde{f}(\vec{v}) = (f(\vec{v}), \ldots, f(\vec{v}))$. Consider a diagonal kernel $\tilde{K} \colon  \mathbb{R}^2 \to \mathbb{R}^{n\times n}$ where $\tilde{K}(\vec{v})$ is a diagonal $n \times n$ matrix $\mathrm{Diag}(K_1,\ldots,K_n)$ and each $K_i\colon \mathbb{R}^2 \to \mathbb{R}^{1\times 1}$}.

 For such inputs and kernels, we have the following permutation equivariance.
     
  \begin{lemma}\label{lem:convequ}   
\begin{align*}
    (\rho_{\mathrm{reg}}(g)\tilde{K}) \star \tilde{f} &= \rho_{\mathrm{reg}}(g)(\tilde{K}\star \tilde{f})
\end{align*}
   \end{lemma}
   
   \begin{proof}
   By definition $h_i = (\tilde{K}\star \tilde{f})_i = K_i \star f$. {Define $h=(h_1,\cdots,h_n)$ and it is clear that permuting the 1x1 kernels $K_i$ also permutes $h_i$, so $\rho_{\mathrm{reg}}(g)h = (\rho_{\mathrm{reg}}(g)\tilde{K})\star \tilde{f}$ as desired.}
   \end{proof}
   
   We require one more lemma on the equivariance of the lifting operator $\mathcal{R}_n$.
   
   \begin{lemma}\label{lem:rnequ}
   \begin{align*}
       \mathcal{R}_n(T_g^0 f) = \rho_{\mathrm{reg}}(-g)\mathcal{R}_n(f)
   \end{align*}
   \end{lemma}
   
   \begin{proof}
   First, we compute 
     \begin{align*}
         \mathcal{R}_n(f)(\vec{x})&=( f(\vec{x}),  f(g^{-1}\vec{x}),\ldots ,f(g^{-(n-1)}\vec{x})).
    \end{align*}
    Then both $\mathcal{R}_n(T_g^0 f)$ and  $\rho_{\mathrm{reg}}(-g)\mathcal{R}_n(f)$ equal to
    \begin{align*}
        (f(g^{-1}\vec{x}),\ldots,f(g^{-(n-1)}\vec{x}), f(\vec{x})).
     \end{align*}
   \end{proof}

\begin{figure}[tp]
    \centering
    \includegraphics[width=0.35\textwidth]{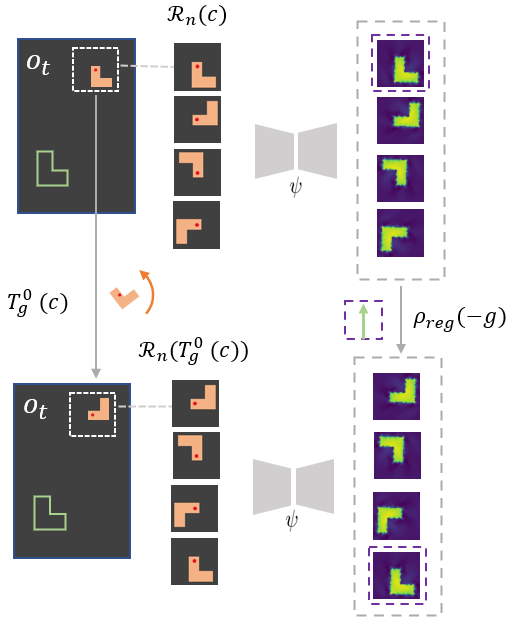}
    \caption{Illustration of the main part of the proof of Proposition~\ref{prop:equivtransporter}.   Rotating the crop $c$ induces a cyclic shift in the channels of the output $\psi(\mathcal{R}_n(T_g^0 c)) = \rho_\mathrm{reg}(-g)\psi(\mathcal{R}_n(c)).$ }
    \label{fig:euqi_transporter}
\end{figure}
\subsubsection{Proof of Proposition 1}
We prove the {$C_n$-place} equivariance of Transporter Net under rotations of the picked object,
  \begin{equation}
    \label{eqn:appprop1}
      \psi(\mathcal{R}_n(T^0_{g}c)) \star \phi(o_t) = \rho_{\reg}(-g)( \psi(\mathcal{R}_n(c)) \star \phi(o_t)
  \end{equation}
  \begin{proof}
   Since $\psi$ is applied independently to each of the rotated channels in $\mathcal{R}_n(c)$, we denote $\psi_n((f_1,\ldots,f_n))=(\psi(f_1),\ldots,(\psi(f_n))$.  By Lemma \ref{lem:rnequ}, the left-hand side of Equation \ref{eqn:appprop1} is  
  \begin{align*}
  \psi(\mathcal{R}_n(T^0_{g}c)) \star \phi(o_t) &= \psi_n(\rho_{\reg}(-g)\mathcal{R}_n(c)) \star \phi(o_t). 
  \end{align*}
  Since $\psi_n$ applies $\psi$ on each component, it is equivariant to the permutation of components and thus the above equation becomes
  \begin{align*}    
       &= (\rho_{\reg}(-g) \psi_n(\mathcal{R}_n(c)) \star \phi(o_t).
  \end{align*}
  Finally applying Lemma \ref{lem:convequ} gives 
  \begin{align*}
      &= \rho_{\reg}(-g) (\psi_n(\mathcal{R}_n(c) \star \phi(o_t))
  \end{align*}
  as desired.
  \end{proof}

The main idea of the proof is shown in Figure \ref{fig:euqi_transporter}.  Namely, $\psi(\mathcal{R}_n(\cdot))$ is equivariant in the sense that rotating the crop $c$ induces a cyclic shift in the channels of the output. Formally, $\psi(\mathcal{R}_n(T_g^0 c)) = \rho_\mathrm{reg}(-g)\psi(\mathcal{R}_n(c)).$  Noting that a permutation of the filters $K$ in the convolution $K \star \phi(o_t)$ induces the same permutation in the output feature maps completes the proof. 
Here $\psi$ is a simple CNN with no rotational equivariance.  The equivariance results from the lifting operator $\mathcal{R}_n$.

 However, only the place network of Transporter Net has the $C_n$-equivariance. Instead, our proposed method incorporates {not only} the rotational equivariance in the pick network {but also}  $C_n\times C_n$-equivariance in the place network.

\section{Equivariant Transporter}

\subsection{Equivariant Pick}

Our approach to the pick network is similar to that in Transporter Net~\cite{zeng2021transporter} except that: 1) we explicitly encode {\emph{the pick symmetry}} into the pick networks, thereby making pick learning more sample efficient; 2) we {consider} the pick orientation so that we can use parallel jaw grippers rather than just suction grippers.

\begin{figure}
    \centering
    \includegraphics[width=0.25\textwidth]{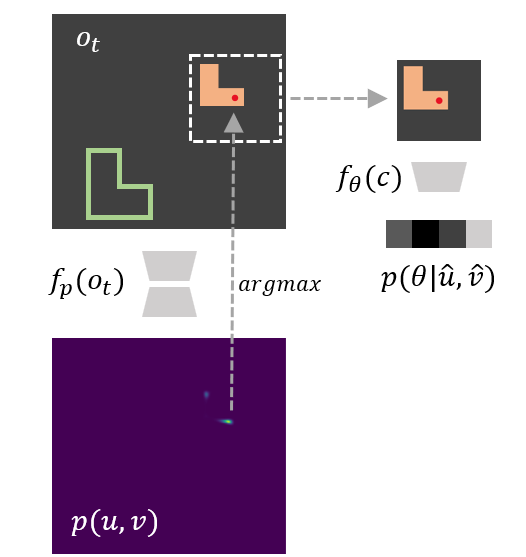}
     
     \caption{Equivariant Transporter Pick model. First, we find the pick position $a^*_{pick}$ by evaluating the argmax over $f_p(o_t)$. Then, we evaluate $f_\theta$ for the image patch centered on $a^*_{pick}$.}
    
    \label{fig:pick_archi}
\end{figure}

\subsubsection{Model}

We propose an equivariant model for detecting the planar pick pose. First, we decompose the learning process of $a_{\pick} \in \SE(2)$ into two parts,
\begin{equation}
    p(a_{\pick}) = p(u,v) p(\theta|(u,v)),
\end{equation}
where $p(u,v)$ denotes {the probability of success when a pick exists at pixel coordinates $u,v$} and $p(\theta|(u,v))$ is the probability that the pick at $u,v$ should be executed with a gripper orientation of $\theta$. The distributions $p(u,v)$ and $p(\theta|(u,v))$ are modeled as two neural networks:
\begin{align}
\label{eqn:pick1}
f_{p}(o_t) &\mapsto p(u,v), \\
\label{eqn:pick2}
f_{\theta}(o_t,(u,v)) &\mapsto p(\theta|(u,v)).
\end{align}

Given this factorization, we can query the maximum of $p(a_{\pick})$ by evaluating $(\hat{u},\hat{v}) = \argmax_{(u,v)}(p(u,v))$ and then $\hat{\theta} = \argmax_{\theta}(p(\theta|\hat{u},\hat{v}))$.
This is illustrated in \figref{fig:pick_archi}. The {bottom} of \figref{fig:pick_archi} shows the maximization of $f_p$ at $a^*_{pick}$. The right side shows the evaluation of $f_\theta$ for the image patch centered at $a^*_{pick}$. 

\subsubsection{{Pick Symmetry}}

There are two equivariant relationships that we would expect to be satisfied for planar picking:
\begin{align}
    \label{eqn:equi1}
    f_p(T^0_g(o_t)) &= T^0_g (f_p(o_t))\\
    \label{eqn:equi2}
    {f_\theta(T^0_g(o_t),T^0_g(u,v))} &{= s(g)(f_{\theta}(o_t,(u,v)))}
\end{align}
{where $s$ is the shift operator and satisfies $s(g)f(x) = f(x+g)$}.

\vspace{0.2cm}
Equation~\ref{eqn:equi1} states that {the pick location distribution} found in an image rotated by $g \in \SO(2)$, (LHS of Equation~\ref{eqn:equi1}), should correspond to {the distribution} found in the original image subsequently rotated by $g$, (RHS of Equation~\ref{eqn:equi1}).

\vspace{0.2cm}
Equation~\ref{eqn:equi2} says that {the pick orientation distribution} at the rotated grasp point $T^0_g(u,v)$ in the rotated image $T^0_g(o_t)$ (LHS of Equation~\ref{eqn:equi2}) should be shifted by $g$ relative to the grasp orientation at the original grasp points in the original image (RHS of Equation~\ref{eqn:equi2}). 

\vspace{0.2cm}
{We encode both $f_p$ and $f_\theta$ using equivariant convolutional layers~\cite{weiler2019general} which constrain the models to represent only those functions that satisfy Equations~\ref{eqn:equi1} and~\ref{eqn:equi2}. Specifically, we select the trivial representation as the output type for $f_p$ and the regular representation as the output type for $f_{\theta}$, which is a \emph{special case}\footnote{{The $\SOtwo$ equivariance can be approximately achieved with irreducible representations as the output type for $f_\theta$.}} of Equation~\ref{eqn:equi2}
\begin{equation}
    f_\theta(T^0_g(o_t),T^0_g(u,v)) = p_{\reg}(g)(f_{\theta}(o_t,(u,v)))\:\:\:\forall g\in C_n
    \label{eqn:special1}
\end{equation}}

\subsubsection{Gripper Orientation Using the Quotient Group}
A key observation in planar picking is that, for many robots, the gripper is bilaterally symmetric, i.e. grasp outcome is invariant when the gripper is rotated by $\pi$. We can encode this additional symmetry to reduce redundancy and save computational cost {using the quotient group $\SOtwo / C_2$ which identifies orientations that are $\pi$ apart.  When using this quotient group for gripper orientation, $s(g)$ in Equation~\ref{eqn:equi2} is replaced with $s(g \:\mathrm{mod}\: \pi)$\footnote{{The $\SOtwo / C_2$ quotient group can be realized by using the basis function with a period of $\pi$.}} and $\rho_{\reg}$ in Equation~\ref{eqn:special1} is replaced with $\rho_{\reg}^{C_n/C_2}$.}

\subsection{Equivariant Place}

{Assumes that the object does not move during picking,} given the picked object represented by the image patch c centered on $a_{\pick}$, the place network models the distribution of $a_{\place}=(u_{\place},v_{\place},\theta_{\place})$ by:
\begin{equation}
    f_{\place}(o_t,c) \mapsto p(a_{\place}|o_t,a_{\pick}),
\end{equation}
where $p(a_{\place}|o_t,a_{\pick})$ denotes the probability that the object at $a_{\pick}$ in scene $o_t$ should be placed at $a_{\place}$.

Our place model architecture closely follows that of Transporter Net~\cite{zeng2021transporter}. The main difference is that we explicitly encode equivariance constraints on both $\phi$ and $\psi$ networks. As a result of this change: 1) we are able to simplify the model by transposing the lifting operation $\mathcal{R}_n$ and the processing by $\phi$; 2) our new model is equivariant with respect to a larger symmetry group $C_n \times C_n$, compared to Transporter Net which is only equivariant over $C_n$.

\subsubsection{Equivariant $\phi$ and $\psi$}
\label{sect:transporter_desc1}

We explicitly encode both $\phi$ and $\psi$ as equivariant models that satisfy the following constraints:
\begin{align}
    \label{eqn:phi}
    \phi(T^0_g(o_t)) &= T^0_g(\phi(o_t))\\
    \label{eqn:psi}
    \psi(T^0_g(c)) &= T^0_g (\psi(c))
\end{align}
for $g \in \SO(2)$.
 The equivariance constraint of Equation~\ref{eqn:phi} says that when the input image rotates, we would expect the place location to rotate correspondingly. This constraint helps the model generalize across place orientations. The constraint of Equation~\ref{eqn:psi} says that when the picked object rotates (represented by the image patch $c$), then the place orientation should correspondingly rotate.

\subsubsection{Place Model}

When the equivariance constraint of Equation~\ref{eqn:psi} is satisfied, we can exchange $\mathcal{R}_n$ (the lifting operation) with $\psi$: 
{
\begin{equation*}
    \psi(\mathcal{R}_n(c)) = \mathcal{R}_n(\psi(c))
\end{equation*}
}
This equality is useful because it means that we only need to evaluate $\psi$ for one image patch and rotate the feature map rather than processing the stack of image patches $\mathcal{R}_n(c)$ -- something that is computationally cheaper. The resulting place model is then:
\begin{eqnarray}
\label{eqn:place_equiv_pre}
f'_{\place}(o_t,c) & = & \mathcal{R}_n(\psi(c)) \star \phi(o_t) \\
\label{eqn:place_equiv}
& = & \Psi(c) \star \phi(o_t),
\end{eqnarray}
where Equation~\ref{eqn:place_equiv} substitutes $\Psi(c) = \mathcal{R}_n(\psi (c))$ to simplify the expression. Here, we use $f'_{\place}$ to denote Equivariant Transporter Net defined using equivariant $\phi$ and $\psi$ in contrast to the baseline Transporter Net $f_{\place}$ of Equation~\ref{eqn:transporter1}. Note that both $f_{\place}$ and $f'_{\place}$ satisfy Proposition~\ref{prop:equivtransporter}. However, $f_{\place}$ accomplishes this by symmetrizing a non-equivariant network (i.e. evaluating $\psi(\mathcal{R}_n(c))$) whereas our model $f'_{\place}$ encodes the symmetry directly into $\psi$.

\begin{figure}[t]
    \centering
    \includegraphics[width=0.40\textwidth]{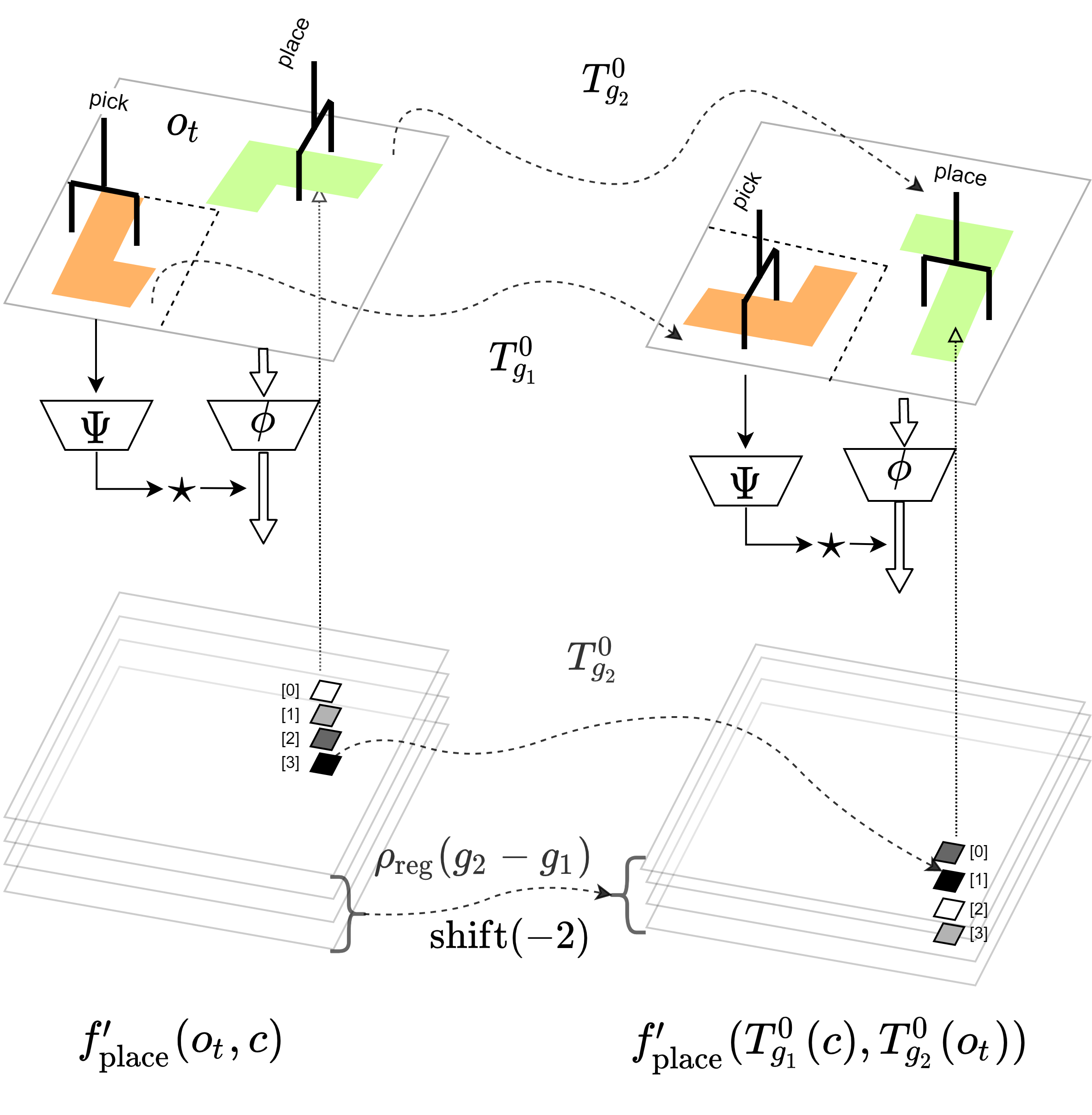}
    \caption[Place Network Architecture]{Equivariance of our placing network under the rotation of the object and the placement. A $\frac{\pi}{2}$ rotation on $c$ and a $-\frac{\pi}{2}$ rotation on $o_t\backslash c$ are equivariant to: i), a $-\frac{\pi}{2}$ rotation on the placing location, and ii), the shift on the channel of placing rotation angle from $\frac{3\pi}{2}$ (the last channel) to $\frac{\pi}{2}$ (the second channel).}
    \label{fig:place_archi}
\end{figure}

\subsection{{Place Symmetry of the Equivariant Transporter Network}}

\subsubsection{{$C_n \times C_n$-place symmetry}}
\label{cncn}
As Proposition~\ref{prop:equivtransporter} demonstrates, the baseline Transporter Net model~\cite{zeng2021transporter} encodes the symmetry that rotations of the object to be picked (represented by $c$) should result in corresponding rotations of the place orientation for that object. However, the pick-conditioned place has a second symmetry that is not encoded in Transporter Net: rotations of the placement (represented by $o_t$) should also result in corresponding rotations of the place orientation. In fact, as we demonstrate in Proposition~\ref{proposition_ours} below, we encode this \emph{second type} of symmetry by enforcing the constraints of Equations~\ref{eqn:phi} and~\ref{eqn:psi}. Essentially, we go from {the $C_n$-place symmetric model to a $C_n \times C_n$-place symmetric model.}

\begin{proposition}
\label{proposition_ours}
Equivariant Transporter Net $f_{\place}'$ is $C_n \times C_n$-equivariant. That is, given rotations $g_1 \in C_n$ of the picked object and $g_2 \in C_n$ of the scene, we have that:

\begin{equation}
        f'_\place(T^0_{g_1}(c), T^0_{g_2} (o_t)) = \rho_{\reg}(g_2-g_1)T^0_{g_2} f'_\place(c, o_t).
    \label{eqn:prop2}
\end{equation}

\end{proposition}
Proposition~\ref{proposition_ours} is illustrated in Figure~\ref{fig:place_archi}. The top of Figure~\ref{fig:place_archi} going left to right shows the rotation of both the object by $g_1$ (in orange) and the place pose by $g_2$ (in green). The LHS of Equation~\ref{eqn:prop2} evaluates $f'_{\place}$ for these two rotated images. The lower left of Figure~\ref{fig:place_archi} shows $f'_{\place}(c,o_t)$. Going left to right at the bottom of Figure~\ref{fig:place_archi} shows the pixel-rotation by $T^0_{g_2}$ and the channel permutation by $g_2 - g_1$ (RHS of Equation~\ref{eqn:prop2}).

To prove Proposition 2, we introduce one more lemma. 

\begin{lemma}
  \begin{align}\label{eqn:lemma1}
    (T_g^0(K\star f))(\vec{v})= ((T_g^0 K) \star (T_g^0 f))(\vec{v})  
  \end{align}
  \end{lemma}
  \noindent
  \begin{proof}
   We evaluate the left-hand side of Equation \ref{eqn:lemma1}:
  \begin{align*}
        T_g^0(K \star f)(\vec{v}) &=  \sum_{\vec{w}\in \mathbb{Z}^2}f(g^{-1}\vec{v}+\vec{w})K(\vec{w }).
  \end{align*}
  Re-indexing the sum with $\vec{y} = g \vec{w}$,
    \begin{align*}
        &= \sum_{\vec{y}\in \mathbb{Z}^2}f(g^{-1}\vec{v}+g^{-1}\vec{y})K(g^{-1}\vec{y })
    \end{align*}
    is by definition
    \begin{align*}
        &= \sum_{\vec{y}\in \mathbb{Z}^2}(T^0_gf)(\vec{v}+\vec{y})(T^0_gK)(\vec{y }) \\
        &= ((T_g^0K)\star (T_g^0f))(\vec{v})
    \end{align*}
    as desired.
  \end{proof}

\subsubsection{Proof of Proposition 2}
Recall $\Psi(c) = \psi(\mathcal{R}_n(c))$. We now prove Proposition 2, 

\begin{align*}    
      \Psi(T^0_{g_1}(c)) \star \phi (T^0_{g_2} (o_t)) =\\ \rho_{reg}(g_2-g_1)(T^0_{g_2}(\Psi(c) \star \phi ( o_t)))
      \label{equ:proof1}
\end{align*}

\begin{proof}
 We first prove the equivariance under rotations of the placement $o_t$.  We claim
\begin{equation}\label{eqn:prop2subclaim}
    \Psi(c) \star \phi (T^0_g(o_t)) = T_g^{\reg}( \Psi(c) \star \phi (o_t)).
  \end{equation}
Evaluating the left hand side of Equation \ref{eqn:prop2subclaim},
 \begin{align*}
    &\Psi(c) \star \phi (T^0_g(o_t)) \\
     &\qquad= \Psi(c) \star T^0_g \phi (o_t) \:\:\:\: \text{(equivariance of $\phi$)}\\
     &\qquad= (T^0_g T^0_{g^{-1}}\Psi(c)) \star (T^0_g \phi (o_t))\\
     &\qquad= T^0_g (T^0_{g^{-1}}\Psi(c) \star \phi (o_t))\:\: \text{(Lemma 3)}\\
     &\qquad= T^0_g (T^0_{g^{-1}}\mathcal{R}_n(\psi (c)) \star \phi (o_t))\\     
     &\qquad= {T^0_g (\mathcal{R}_n(T^0_{g^{-1}}\psi (c)) \star \phi (o_t))} \:\: 
     \text{{(equivariance of $\mathcal{R}_n$)}}\\
     &\qquad= {T^0_g (\mathcal{R}_n(\psi (T^0_{g^{-1}} c)) \star \phi (o_t))} \:\: \text{{(equivariance of $\psi$)}}\\
     &\qquad= T^0_g ((\rho_{\reg}(g)\Psi(c) \star \phi (o_t)) \:\:\text{(Lemma 2)}\\
     &\qquad= T^0_g \rho_{\reg}(g)(\Psi(c) \star \phi (o_t)) \:\: \text{(Lemma 1)}\\
     &\qquad=  T_g^{\reg}( \Psi(c) \star \phi (o_t)).
 \end{align*}
 In the last step, $T_g^{\reg} = \rho_{\reg}(g) T^0_g  = T^0_g \rho_{\reg}(g)$ since  $T^0_g$ and $\rho_{\reg}(g)$ commute as $\rho_{\reg}(g)$ acts on the channel space and $T_g^0$ acts on the base space.  This proves the claim of Equation \ref{eqn:prop2subclaim}.
 
{Recall $\Psi(c) = \mathcal{R}_n(\psi (c))$.} Using the equivariance of $\psi$, Proposition~\ref{prop:equivtransporter} could be reformulated as
 \begin{equation}
      \Psi(T^0_{g}c) \star \phi(o_t) = \rho_{\reg}(-g)( \Psi(c) \star \phi(o_t))
      \label{equ:proof2}
 \end{equation}
 Evaluating the left hand side of Equation~\ref{equ:proof2},
\begin{align*}
    &\Psi(T^0_{g}c) \star \phi(o_t)\\
    &\qquad =\mathcal{R}_n(\psi(T^0_g c))\star \phi(o_t)\:\:\text{($\Psi(c) = \mathcal{R}_n(\psi(c))$ by def.)}\\
    &\qquad =\psi(\mathcal{R}_n(T^0_g c))\star \phi(o_t) \:\:\text{(equivariance of $\psi$)}\\
    &\qquad = \rho_{\reg}(-g)( \psi(\mathcal{R}_n(c)) \star \phi(o_t))\:\:\text{(Proposition 1)}\\
    &\qquad = \rho_{\reg}(-g)( \mathcal{R}_n(\psi(c)) \star \phi(o_t))\:\:\text{(equivariance of $\psi$)}\\
    &\qquad = \rho_{\reg}(-g)( \Psi(c) \star \phi(o_t))
\end{align*}
 
 Combining \eref{eqn:prop2subclaim} with \eref{equ:proof2} realizes the Proposition 2. 
\end{proof}

\subsubsection{Translational Symmetry}
Note that in addition to the two rotational symmetries enforced by our model, it also has translational symmetry.  Since the rotational symmetry is realized by additional restrictions to the weights of kernels of convolutional networks, the rotational symmetry is in addition to the underlying shift equivariance of the convolutional network. Thus, the full symmetry group enforced is the group generated by $C_n \times C_n \times (\mathbb{R}^2,+)$.
Equivariant neural networks learn effectively on a lower dimensional space, the equivalence classes of samples under the group action. 

\subsubsection{{From $C_n\times C_n$-place symmetry to $\SOtwo \times \SOtwo$}}
    {The above place symmetry is limited to the cyclic group due to the role of $\mathcal{R}_n$, though as $n\rightarrow \infty$, $C_n$ equals $\SOtwo$. We show the generalization of the $C_n\times C_n$-place symmetry and $\SOtwo\times \SOtwo$ place symmetry below.}

\begin{proposition}
    \label{pro:steerable}
        {Given $g \in G$ for $G\subseteq \SOtwo$, an equivariant model $\phi$ satisfying $\phi(T^0_g(o_t))= T^0_g(\phi(o_t))$ and a function $\Bar{\Psi}\colon c \mapsto K$ satisfying the equivariant constraint
        $ \Bar{\Psi}(T_g^0c) =  T_g^0\Bar{\Psi}(c)$, where $c$ is the crop $\in \mathbb{R}^2$ and $K \colon \mathbb{R}^{2} \rightarrow \mathbb{R}^{d_{\mathrm{out}}\times d_{\mathrm{trivial}}}$ is a 2D steerable kernel
        with {trivial representation} as the input type.
        The cross-correlation between $\Bar{\Psi}(c)$ and $\phi(T^0_g(o_t))$ satisfies}
    \begin{align}
    \label{equ:generalized}
        {\Bar{\Psi}(T^0_{g_1}(c)) \star \phi (T^0_{g_2} (o_t))=\rho_{\mathrm{out}}(g_2-g_1)(T^0_{g_2}(\Bar{\Psi}(c) \star \phi ( o_t)))}
    \end{align}
\end{proposition}

{Proposition~\ref{pro:steerable} states that to satisfy the cross-type place symmetry, one necessary condition is that the output of $\Bar{\Psi}$ is a steerable kernel. It generalizes Proposition~\ref{proposition_ours} to either $C_n$ or $\SOtwo$. In fact, $\Psi(c) = \mathcal{R}_n(\psi (c))$ combining the lift operator $\mathcal{R}_n$ and the equivariant constraint of $\psi$ shown in Equation~\ref{eqn:psi} is a special case of $\Bar{\Psi}(c)$. $\mathcal{R}_n \colon \mathbb{R}^2 \rightarrow K$ outputting a steerable kernel $K$ that takes \emph{the regular representation} of $C_n$ as the output type and satisfies the steerability constraint of Equation~\ref{equ:steerablility_constraint}. When using \emph{irreducible representations} as the output type, we can instantiate $\rho_{\mathrm{out}}(g_2-g_1)$ in RHS of Equation~\ref{equ:generalized} as $\rho_{\mathrm{irrep}}(g_2-g_1)$ which is equivalent to $s(g_2 - g_1)$ after Inverse Fourier Transform.}

{To prove Proposition~\ref{pro:steerable}, we first introduce another lemma.}
\begin{lemma}
\label{lemma4}
{A 2D steerable kernel $K \colon \mathbb{R}^{2} \rightarrow \mathbb{R}^{d_{\mathrm{out}}\times d_{\mathrm{trivial}}}$ satisfies}
\begin{equation}
    {T_g^{0}K(x) =  \rho_{\mathrm{out}}(g^{-1})K(x)}
\end{equation}
\end{lemma}

\begin{proof}
    {Recall that $\rho_{0}(g)$ is an identity mapping. Substituting $\rho_{\mathrm{in}}$ with  $\rho_{0}(g)$ and $g^{-1}$ with $g$ in the steerability constraint
    $K(g\cdot x) = \rho_{\mathrm{out}}(g)K(x)\rho_{\mathrm{in}}(g)^{-1}$ 
    shown in~\eref{equ:steerablility_constraint}
    completes the proof.}
    \begin{align*}
        T_g^0K(x) &= K(g^{-1} x)\\
        &=\rho_{\mathrm{out}}(g^{-1})K(x)\rho_{\mathrm{in}}(g)\\
        &=\rho_{\mathrm{out}}(g^{-1})K(x)
    \end{align*}

\end{proof}

{Lemma~\ref{lemma4} states that when the input type is the trivial representation, a spatial rotation of the steerable kernel is the same as the inverse channel space transformation. With Lemma~\ref{lemma4} in hand, we start the proof of proposition~\ref{pro:steerable}}
\begin{proof}
 {Similar to the proof of Proposition~\ref{proposition_ours}, we first show the equivariance under rotations of the placement $o_t$.  We claim}
\begin{equation}\label{eqn:prop3subclaim}
    {\Bar{\Psi}(c) \star \phi (T^0_g(o_t)) = T_g^{\mathrm{out}}( \Bar{\Psi}(c) \star \phi (o_t))}
\end{equation}
 {Starting from the left-hand side of Equation \ref{eqn:prop3subclaim},}
 \begin{align*}
     &\Bar{\Psi}(c) \star \phi (T^0_g(o_t)) \\
     &\qquad= \Bar{\Psi}(c) \star T^0_g \phi (o_t) \:\:\:\: \text{(equivariance of $\phi$)}\\
     &\qquad= (T^0_g T^0_{g^{-1}}\Bar{\Psi}(c)) \star (T^0_g \phi (o_t))\\
     &\qquad= T^0_g (T^0_{g^{-1}}\Bar{\Psi}(c) \star \phi (o_t))\:\: \text{(Lemma 3)}\\
     &\qquad= T^0_g (\rho_{\mathrm{out}}(g)\Bar{\Psi} (c) \star \phi (o_t)) \:\:\text{(Lemma 4)}\\     
     &\qquad=  T_g^{\mathrm{out}}( \Bar{\Psi}(c) \star \phi (o_t)).
 \end{align*}

 {Then, we propose the equivariance under rotations of the picked object as}
 \begin{equation}\label{eqn:prop3subclaim2}
     \Bar{\Psi}(T^0_g(c)) \star \phi (o_t) = \rho_{\mathrm{out}}(-g)(\Bar{\Psi}(c) \star \phi (o_t))
 \end{equation}

 {Evaluating the left-hand side of Equation~\ref{eqn:prop3subclaim2}},
  \begin{align*}
    &\Bar{\Psi}(T^0_g(c)) \star \phi (o_t) \\
     &\qquad= T^0_g\Bar{\Psi}(c) \star \phi (o_t) \:\:\:\: \text{(equivariance of $\Bar{\Psi}$)}\\
     &\qquad= \rho_{\mathrm{out}}(-g)\Bar{\Psi }(c) \star \phi (o_t)\:\: \text{(Lemma 4)}
 \end{align*}
 {Combining Equation~\ref{eqn:prop3subclaim} with Equation~\ref{eqn:prop3subclaim2} realizes the Proposition 3.}
  \end{proof}
  
  {Note that Proposition~\ref{pro:steerable} gives the way to realize the $\SOtwo$ version of our model 
  and provide some insights to extend it to 3D signals without limitations. {That is generating the 3D dynamic steerable kernels from the crop signal.}
  But in this work, we primarily focus on the discrete $C_n$ group since it is easy to compare with the baseline \emph{Transporter Net} on Ravens-10 Benchmark. 
  }
\subsection{Analyzing equivariance under Proposition 2}
\label{special_properties}
We summarize some important properties from {the larger symmetry group of our place network} and provide an intuitive explanation for each one. Recall that Proposition 2 states:
\begin{equation*}
    \begin{aligned}
        &\Psi(T^0_{g_1}(c)) \star \phi (T^0_{g_2} (o_t)) \\ &\qquad=\rho_{\reg}(g_2-g_1)(T^0_{g_2}(\Psi(c) \star \phi ( o_t))).
    \end{aligned}
    \label{generalized}
\end{equation*}
Then we have the following properties:

\paragraph{Equivariance property}
Setting {either} $g_1=0$ or $g_2=0$ we get respectively
\begin{align}
    \Psi(T^0_g(c)) \star \phi (o_t) &= \rho_{\reg}(-g) (\Psi(c) \star \phi (o_t))
    \label{equi_original_our} \\
    \Psi(c) \star \phi (T^0_g(o_t)) &= T^{\reg}_g(\Psi(c) \star \phi (o_t))
    \label{equi_2_simple}
\end{align}
These show the equivariance of our network $f_\place$ under either a rotation $g\in C_n$ of the object or the placement.

\paragraph{Invariance property} Setting $g_1=g_2$, we get
\begin{equation}
    \Psi(T^0_g(c)) \star \phi (T^0_g( o_t)) = T^0_g (\Psi(c) \star \phi (o_t)).
\end{equation}
This equation demonstrates that a rotation $g$ on the whole observation $o_t$ {including the objects} does not change the placing angle but rotates the placing location by $g$. Although data augmentation could help non-equivariant models learn this property, our networks observe it by construction. {Note that for the discrete group, data augmentation propagates to every element within the group.}

\paragraph{Relativity property}
Related to Equation \ref{equi_original_our}, we also have
\begin{equation*}
    \begin{aligned}
        \Psi(T^0_g(c)) \star \phi (o_t) =  \rho_{reg}(-g)(T^0_g( \Psi(c) \star \phi (T^0_{-g}(o_t))))
    \end{aligned}
    \label{relativity}
\end{equation*}
This equation defines the \emph{dual} relationship between a rotation on $c$ by $g$ and an inverse rotation $-g$ on $o_t$. Intuitively, $c$ could be considered as the L-shaped block and $o_t$ can be regarded as the L-shaped slot. A rotation on the picked object is equivariant to an inverse rotation on the placement under some transformation.

\subsection{Goal-conditioned equivariant transporter network}
{The goal-conditioned pick-place task is an important branch in learning manipulation skills where the goal could be represented as language instructions, images, or other customized definitions.}
\cite{seita_bags_2021} extended Transporter Net to solve image-based goal-conditioned tasks. In this setting, the goal is represented explicitly as an image that is part of the problem input rather than implicitly as part of the observation. 
Two goal-conditioned architectures are proposed in~\cite{seita_bags_2021}. \textit{Transporter-Goal-Stack}
stacks the current $o_t$ and goal $o_g$ images channel-wise and passes it as input through a standard Transporter Network. \textit{Transporter-Goal-Split} processes
the goal image $o_g$ through a separate Fully Convolution Network $\phi_{\mathrm{goal}}$ to generate dense features {to be combined with dense features of $\phi_{\mathrm{query}}{(o_t)}$ using the Hadamard product to infer the goal-conditioned pick}
\begin{align}
    f_{\mathrm{query}} &= \phi_{\mathrm{query}}(o_t) \odot \phi_{\mathrm{goal}}(o_g)
\end{align}
{and evaluate the goal-conditioned place with}
\begin{align}
    f_{\mathrm{key}} &= \phi_{\mathrm{key}}(o_t) \odot \phi_{\mathrm{goal}}(o_g)\\
    p(a_{\place}|o_t,o_g,a_{\pick}) &= \mathcal{R}_n(f_{\mathrm{query}}[a_\mathrm{pick}]) \star f_{\mathrm{key}}
\end{align}
where $f_{\mathrm{query}}[a_\mathrm{pick}]$ denotes the crop of the dense feature map $f_{\mathrm{query}}$ centered on $a_{\mathrm{pick}}$ and $\odot$ is the the Hadamard product.

{Since the pick-place symmetries also exist in goal-conditioned tasks, we realize the goal-conditioned equivariant transporter with some simple modifications.} Denote $||$ as the channel-wise concatenation, the $C_n$-equivariance of the picking network holds when stacking $o_t$ and $o_g$ as the input:
\begin{equation}
    f_{\mathrm{p}}(T^0_{g} (o_t||o_g)) = T^0_g  f_{\mathrm{p}}(o_t||o_g)
    \label{euq:goal_pick}
\end{equation}
The $C_n\times C_n$-equivariance of the placing model also holds:
\begin{multline}    
      \Psi(T^0_{g_1}((o_t||o_g)[a_\mathrm{pick}])) \star \phi (T^0_{g_2} (o_t||o_g)) =\\ \rho_{reg}(g_2-g_1)(T^0_{g_2}(\Psi((o_t||o_g)(a_\mathrm{pick})) \star \phi ( o_t||o_g)))
      \label{euq:goal_place}
\end{multline}

Based on the two equations above, we define \textit{Equivariant-Transporter-Goal-Stack} to solve goal-conditioned tasks.

\begin{figure*}[ht]
    \centering
    \includegraphics[width=1\textwidth]{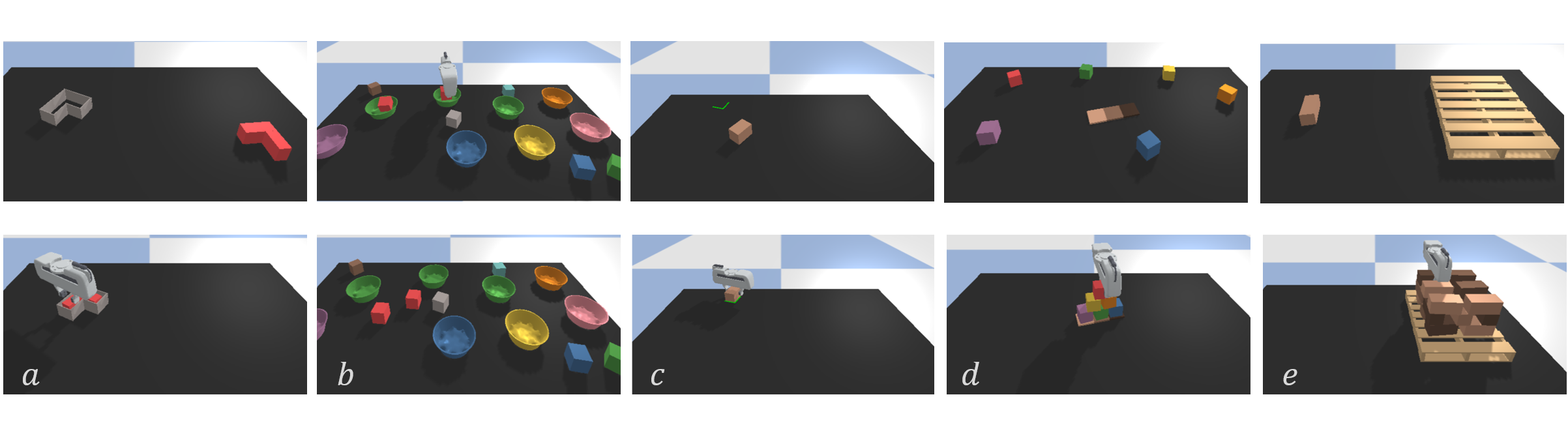}
    \caption{ Simulated environment for parallel-jaw gripper tasks. From left to right: (a) inserting blocks into fixtures, (b) placing red boxes into green bowls, (c) align box corners to green lines, (d) stacking a pyramid of blocks, (e) palletizing boxes.}
    \label{fig:gripper_env}
\end{figure*}

\subsection{Model Architecture Details}
\subsubsection{Pick model $f_p$ (Equation~\ref{eqn:pick1})}

The input to $f_p$ is a $4$-channel RGB-D image $o_t \in \mathbb{R}^{4\times H \times W}$. The output is a feature map $p(u,v) \in \mathbb{R}^{H\times W}$ which encodes a distribution over pick location. $f_p$ is implemented as an 18-layer equivariant residual network with a U-Net~\cite{ronneberger2015u} as the main block. The U-net has 8 residual blocks (each block contains 2 equivariant convolution layers~\cite{weiler2019general} and one skip connection): 4 residual blocks~\cite{he2016deep} are used for the encoder and the other 4 residual blocks are used for the decoder. The encoding process trades spatial dimensions for channels with max-pooling in each block; the decoding process upsamples the feature embedding with bilinear-upsampling operations. The first layer maps the trivial representation of $o_t$ to regular representation and the last equivariant layer transforms the regular representation back to the trivial representation, followed by image-wide softmax. ReLU activations~\cite{nair2010rectified} are interleaved inside the network. 

\subsubsection{Pick model $f_\theta$ (Equation~\ref{eqn:pick2})}

Given the picking location $(u^*,v^*)$, the pick angle network $f_\theta$ takes as input a crop $c\in \mathbb{R}^{4\times H_1 \times W_1}$ centered on $(u^*,v^*)$ and outputs the distribution $p(\theta|u,v) \in \mathbb{R}^{n/2}$, where $n$ is the size of the rotation group (i.e. $n=| C_n |$). The first layer maps the trivial representation of $c$ to a quotient regular representation followed by 3 residual blocks containing max-pooling operators. This goes to two equivariant convolution layers and then to an average pooling layer.

\subsubsection{Place models $\phi$ and $\psi$}

Our place model has two equivariant convolution networks, $\phi$ and $\psi$, and both have similar architectures to $f_p$. The network $\phi$ takes as input a zero-padded version of the 4-channel RGBD observation $o_t$, $\mathrm{pad}(o_t)\in \mathbb{R}^{4\times (H+d) \times (W+d)}$, and generates a dense feature map, $\phi(\mathrm{pad}(o_t))\in\mathbb{R}^{(H+d) \times (W+d)}$, where $d$ is the padding size. The network $\psi$ takes as input the image patch $c \in \mathbb{R}^{4\times H_2 \times W_2}$ and outputs $\psi(c)\in \mathbb{R}^{H_2 \times W_2}$. After applying rotations of $C_n$ to $\psi(c)$, the transformed dense embeddings $\Psi(c)\in \mathbb{R}^{n\times H_2 \times W_2}$ are cross-correlated with $\phi(\mathrm{pad}(o_t))$ to generate the placing action distribution $p(a_{\place}|o_t,a_{\pick}) \in \mathbb{R}^{n\times H \times W}$, where the channel axis $n$ corresponds to placing angles, $\frac{2\pi i}{n}$ for $0\leq i < n$.

\subsubsection{{Group Types and Sizes}}

{
The networks $f_p$, $\psi$, and $\phi$ $\colon \rho_{0}\mapsto \rho_{0}$, which are all defined using $C_6$ regular representations in the intermediate layers. The ablation study of the group size of the latent feature is discussed in the Experiment section. The network $f_\theta \colon \rho_{0}\mapsto \rho_{\mathrm{quot}}$ is defined using the quotient representation $C_{36}/C_2$, which corresponds to the number of allowed pick orientations. The lift operator $\mathcal{R}_n$ is implemented with $C_{36}$ cyclic group, which allows 36 different place orientations. {Both the number of allowed pick and place orientations are hyperparameters and could be selected flexibly based on the task precision}. Our choice of the $\frac{\pi}{18}$ discretization, i.e., 18 bilateral-symmetric pick orientations and 36 place orientations, follows the settings of Ravens-10 benchmark~\cite{zeng2021transporter}.}

\subsubsection{Training Details}
\label{traing_settings}

We train Equivariant Transporter Network with the Adam~\cite{kingma2014adam} optimizer with a fixed learning rate of $10^{-4}$. It takes about 0.8 seconds\footnote{{The resolution of the input image is ${320\times 160}$ for this training time.}} to complete one SGD step with a batch size of one on an NVIDIA Tesla V100 SXM2 GPU. {Compared with the baseline transporter net which takes around 0.6 seconds to complete one SGD step on the same setting, the equivariant constraint on the weight updating increases $33\%$ computation load.} {In fact, Equivariant Transporter Net converges faster than the baseline Transporter Net as shown in Figure~\ref{fig:fast_converge}. This is due to that the larger symmetry group results in a smaller dimensional sample space and thus better coverage by the training data.}
{For each task, we train a single-policy network and evaluate the performance every 1k steps on 100 unseen tests.} On most tasks, the best performance is achieved in less than 10k SGD steps.

\section{Experiments}

We evaluate Equivariant Transporter using the Ravens-10 Benchmark~\cite{zeng2021transporter} and our variations thereof.

\subsection{Tasks}

\subsubsection{Ravens-10 Tasks}

Ravens-10 is a behaviour cloning simulation benchmark for manipulation, where each task owns an oracle that can sample expert demonstrations from the distribution of successful picking and placing actions with access to the ground-truth pose of each object. The 10 tasks of Ravens can be classified into 3 categories: \textit{Single-object manipulation tasks} (block-insertion, align-box-corner); \textit{Multiple-object manipulation tasks} (place-red-in-green, towers-of-hanoi, stack-block-pyramid, palletizing-boxes, assembling-kits, packing-boxes); \textit{Deformable-object manipulation task} (manipulating-rope, sweeping-piles).

Here we provide a short description of Ravens-10 Environment, we refer readers to~\cite{zeng2021transporter} for details.
The poses of objects and placements in each task are randomly sampled in the workspace without collision. Performance on each task is evaluated in one of two ways: 1) pose: translation and rotation error relative to target pose is less than a threshold $ \tau=1\mathrm{cm}$ and $\omega=\frac{\pi}{12}$ respectively. Tasks: block-insertion, towers-of-hanoi, place-red-in-green, align-box-corner, stack-block-pyramid, assembling-kits. Partial scores are assigned to multiple-action tasks. 2) Zone:  Ravens-10 discretizes the 3D bounding box of each object into $2cm^3$ voxels. The Total reward is calculated by $\frac{\text{\# of voxels in target zone}}{\text{total \# of voxels}}$. Tasks: palletizing-boxes, packing-boxes, manipulating-cables, sweeping-piles.
Note that pushing objects could also be parameterized with $a_{\mathrm{pick}}$ and $a_\mathrm{{place}}$ that correspond to the starting pose and the ending pose of the end effector.

\begin{figure*}[ht]
     \centering
     \begin{subfigure}[b]{0.245\textwidth}
         \centering
         \includegraphics[width=0.98\textwidth]{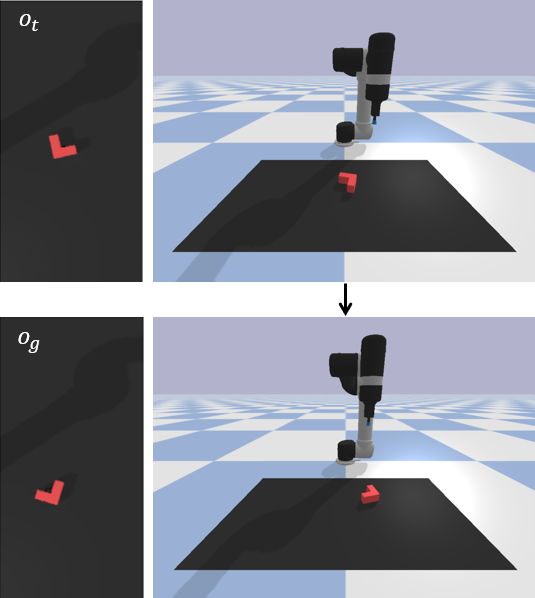}
         \caption{goal-conditioned block insertion}
         \label{fig:block_insertion_goal}
     \end{subfigure}
     \hfill
     \begin{subfigure}[b]{0.245\textwidth}
         \centering
         \includegraphics[width=0.99\textwidth]{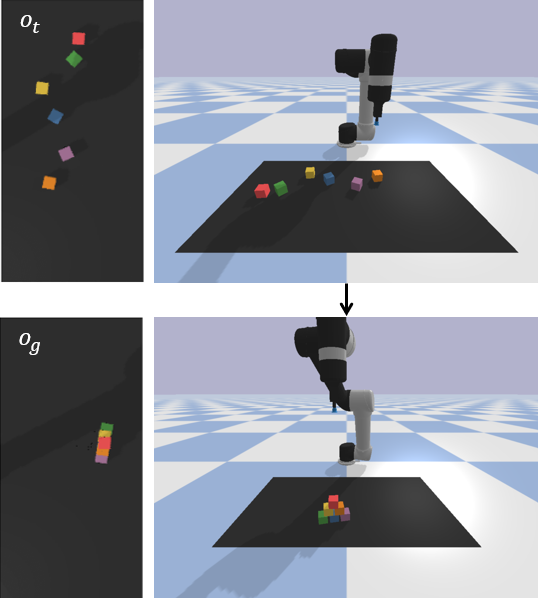}
         \caption{goal-conditioned block pyramid}
         \label{fig:stack_pyramid_goal}
     \end{subfigure}
     \hfill
     \begin{subfigure}[b]{0.245\textwidth}
         \centering
         \includegraphics[width=0.98\textwidth]{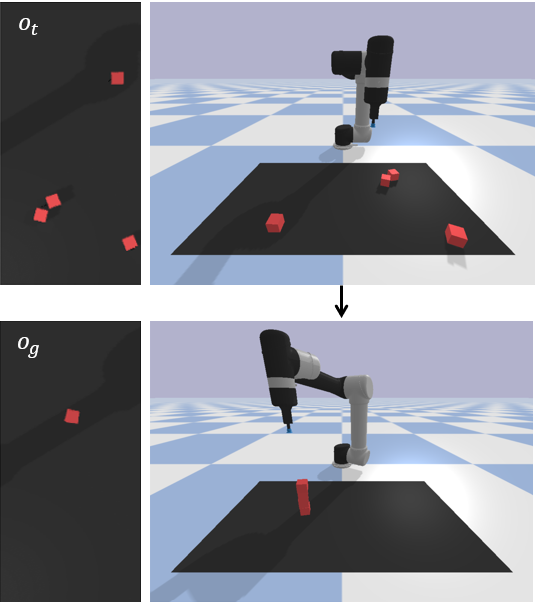}
         \caption{goal-conditioned four blocks}
         \label{fig:stack_4_blocks_goal}
     \end{subfigure}
     \hfill
     \begin{subfigure}[b]{0.245\textwidth}
         \centering
         \includegraphics[width=1.\textwidth]{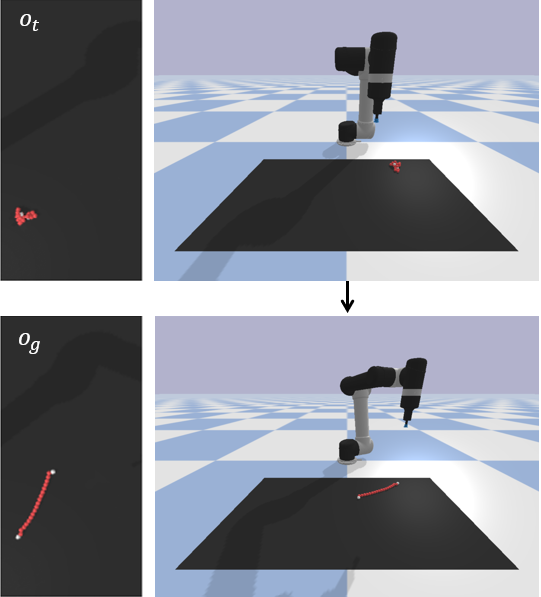}
         \caption{goal-conditioned cable align}
         \label{fig:cable_align_goal}
     \end{subfigure}
     \caption{ Simulated environment for goal-conditioned tasks. The first row shows the initial observations of four tasks where objects are generated with random poses on the workspace. The robot must use pick-and-place actions to manipulate the objects to the target pose specified in the goal images $o_g$, as shown in the second row.}
     \label{fig:goal-conditioned_task}
\end{figure*}

\begin{enumerate}[topsep=0pt,itemsep=-1ex,partopsep=1ex,parsep=1ex]
    \item \textbf{block-insertion:} Picking up an L-shape block and placing it into an L-shaped fixture. 
    \item \textbf{place-red-in-green:} picking up red cubes and placing them into green bowls. There could be multiple bowls and cubes with different colors.
    \item \textbf{towers-of-hanoi:} sequentially picking up disks and placing them into pegs such that all 3 disks initialized on one peg are moved to another, and that only smaller disks can be on top of larger ones.
    \item \textbf{align-box-corner:} picking up {a randomly sized box} and placing it to align one of its
    corners to a green L-shaped marker labeled on the tabletop. {This task requires precision and generalization ability to new box sizes.}
    \item \textbf{stack-block-pyramid:} sequentially picking up 6 blocks and stacking them into a pyramid of 3-2-1.
    \item \textbf{palletizing-boxes:} picking up 18 boxes and stacking them on top of a pallet.
    \item \textbf{assembling-kits:} picking 5 shaped objects ({randomly sampled with replacement from a set of 20 objects, where 14 objects are used during training and 6 objects are held
out for testing}) and fitting them to corresponding silhouettes of the objects on a board. {This task requires generalizing to new objects.}
    \item \textbf{packing-boxes:} picking and placing randomly sized boxes tightly into a randomly sized container.
    \item \textbf{manipulating-rope:} manipulating a deformable rope such that it connects the two endpoints of an incomplete 3-sided square (colored in green).
    \item \textbf{sweeping-piles:} pushing piles of small objects (randomly initialized) into a desired target goal zone on the tabletop marked with green boundaries. The task is implemented with a pad-shaped end effector. 
\end{enumerate}

\subsubsection{Ravens-10 Tasks Modified for the Parallel Jaw Gripper}
\label{parallel env}

We select 5 tasks (block-insertion, align-box-corner, place-red-in-green, stack-block-pyramid, palletizing-boxes) from Ravens-10 and replaced the suction cup with the Franka Emika gripper, which requires additional picking angle inference. \figref{fig:gripper_env} illustrates the initial state and completion state for each of these five tasks. For each of these five tasks, we defined an oracle agent. Since the Transporter Net framework assumes that the object does not move during picking, we defined these expert generators such that this was the case.

\subsubsection{Goal-conditioned tasks}
We design four image-based goal-conditioned tasks (goal-conditioned block insertion, goal-conditioned block pyramid, goal-conditioned four blocks-no, goal-conditioned cable align) based on ravens-10, as shown in ~\figref{fig:goal-conditioned_task}.

For each of the four tasks, objects are generated with random poses on the workspace and there is no placement in the observation $o_t$. The robot must use pick-place actions to manipulate the objects to the target pose specified in the goal images. For the goal-conditioned cable-align task, the robot needs to align the rope to the straight line shown in the goal image.

\begin{table*}[ht]
  \setlength\tabcolsep{4pt}
  \begin{center}
  \scriptsize
  \begin{tabular}{@{}lcccccccccccccccccccc@{}}
  \toprule
  & \multicolumn{4}{c}{block-insertion} & \multicolumn{4}{c}{place-red-in-green} & \multicolumn{4}{c}{towers-of-hanoi} & \multicolumn{4}{c}{align-box-corner} & \multicolumn{4}{c}{stack-block-pyramid} \\
  \cmidrule(lr){2-5} \cmidrule(lr){6-9} \cmidrule(lr){10-13} \cmidrule(lr){14-17} \cmidrule(lr){18-21}
  Method & 1 & 10 & 100 & 1000 & 1 & 10 & 100 & 1000 & 1 & 10 & 100 & 1000 & 1 & 10 & 100 & 1000 & 1 & 10 & 100 & 1000 \\
  \midrule
  Equivariant Transporter                & \textbf{100} & \textbf{100} & \textbf{100} & \textbf{100} & \textbf{98.5} & \textbf{100} & \textbf{100} & \textbf{100} & \textbf{88.1} & \textbf{95.7} & \textbf{100} & \textbf{100} & 41.0 & \textbf{99.0} & \textbf{100} & \textbf{100} & \textbf{34.6} & \textbf{80.0} & \textbf{90.8} & \textbf{95.1} \\
  Transporter Network                  & \textbf{100} & \textbf{100} & \textbf{100} & \textbf{100} &  84.5 & \textbf{100} & \textbf{100} & \textbf{100} & 73.1 & 83.9 & 97.3 & 98.1 & 35.0 & 85.0 & 97.0 & 98.0 & 13.3 & 42.6 & 56.2 & 78.2 \\
  Form2Fit       & 17.0 & 19.0 & 23.0 & 29.0 & 83.4 & \textbf{100} & \textbf{100} & \textbf{100} & 3.6 & 4.4 & 3.7 & 7.0 & 7.0 & 2.0 & 5.0 & 16.0 & 19.7 & 17.5 & 18.5 & 32.5 \\
  Conv. MLP                           & 0.0 & 5.0 & 6.0 & 8.0 & 0.0 & 3.0 & 25.5 & 31.3 & 0.0 & 1.0 & 1.9 & 2.1 & 0.0 & 2.0 & 1.0 & 1.0 & 0.0 & 1.8 & 1.7 & 1.7 \\
  GT-State MLP                        & 4.0 & 52.0 & 96.0 & 99.0 & 0.0 & 0.0 & 3.0 & 82.2 & 10.7 & 10.7 & 6.1 & 5.3 & 47.0 & 29.0 & 29.0 & 59.0 & 0.0 & 0.2 & 1.3 & 15.3 \\
  GT-State MLP 2-Step                        & 6.0 & 38.0 & 95.0 & \textbf{100} & 0.0 & 0.0 & 19.0 & 92.8 & 22.0 & 6.4 & 5.6 & 3.1 & \textbf{49.0} & 12.0 & 43.0 & 55.0 & 0.0 & 0.8 & 12.2 & 17.5 \\
  \midrule
  & \multicolumn{4}{c}{palletizing-boxes} & \multicolumn{4}{c}{assembling-kits} & \multicolumn{4}{c}{packing-boxes} & \multicolumn{4}{c}{manipulating-rope} & \multicolumn{4}{c}{sweeping-piles}\\
  \cmidrule(lr){2-5} \cmidrule(lr){6-9} \cmidrule(lr){10-13} \cmidrule(lr){14-17} \cmidrule(lr){18-21}
  & 1 & 10 & 100 & 1000 & 1 & 10 & 100 & 1000 & 1 & 10 & 100 & 1000 & 1 & 10 & 100 & 1000 & 1 & 10 & 100 & 1000 \\
  \midrule
  Equivariant Transporter                & \textbf{75.3} & \textbf{98.9} & \textbf{{99.6}} & \textbf{{99.6}} & \textbf{63.8} & \textbf{90.6} & \textbf{{98.6}} & \textbf{{100}} & \textbf{{98.3}} & \textbf{{99.4}} & \textbf{{99.6}} & \textbf{{100}} & \textbf{31.0} & \textbf{85.0} & \textbf{92.3} & \textbf{98.4} & \textbf{97.9} & \textbf{99.5} & \textbf{100} & \textbf{100} \\
  Transporter Network                 & 63.2 & 77.4 & 91.7 & 97.9 & 28.4 & 78.6& 90.4& 94.6 & 56.8 & 58.3 & 72.1 & 81.3 & 21.9 & 73.2 & 85.4 & 92.1 & 52.4 & 74.4 & 71.5 & 96.1 \\
  Form2Fit                            & 21.6 & 42.0 & 52.1 & 65.3 & 3.4  & 7.6 & 24.2& 37.6 & 29.9 & 52.5 & 62.3 & 66.8 & 11.9 & 38.8 & 36.7 & 47.7 & 13.2 & 15.6 & 26.7 & 38.4 \\
  Conv. MLP                           & 31.4 & 37.4 & 34.6 & 32.0 & 0.0  & 0.2 & 0.2 & 0.0 & 0.3 & 9.5 & 12.6 & 16.1 & 3.7 & 6.6 & 3.8 & 10.8 & 28.2 & 48.4 & 44.9 & 45.1 \\
  GT-State MLP                        & 0.6  & 6.4  & 30.2 & 30.1 & 0.0  & 0.0 & 1.2 & 11.8 & 7.1 & 1.4 & 33.6 & 56.0 & 5.5 & 11.5 & 43.6 & 47.4 & 7.2 & 20.6 & 63.2 & 74.4 \\
  GT-State MLP 2-Step                 & 0.6  & 9.6  & 32.8 & 37.5 & 0.0  & 0.0 & 1.6 & 4.4 & 4.0 & 3.5 & 43.4 & 57.1 & 6.0 & 8.2 & 41.5 & 58.7 & 9.7 & 21.4 & 66.2 & 73.9 \\
  \bottomrule
  \end{tabular}
  \end{center}
  \vspace{0.5em}
  \caption{\scriptsize\textbf{Performance comparisons on Ravens-10 benchmark (suction gripper).} Success rate (mean\%) vs. the number of demonstration episodes (1, 10, 100, or 1000) used in training. Best performances are highlighted in bold.}
  \vspace{-1.0em}
  \label{table:sample-efficiency1}
\end{table*}

\begin{table*}[ht]
  \setlength\tabcolsep{4pt}
  \centering
  \scriptsize
  \begin{tabular}{@{}lccccccccccccccc@{}}
  \toprule
  & \multicolumn{3}{c}{block-insertion} & \multicolumn{3}{c}{place-red-in-green} & \multicolumn{3}{c}{palletizing-boxes} & \multicolumn{3}{c}{align-box-corner} & \multicolumn{3}{c}{stack-block-pyramid} \\
  \cmidrule(lr){2-4} \cmidrule(lr){5-7} \cmidrule(lr){8-10} \cmidrule(lr){11-13} \cmidrule(lr){14-16}
  Method & {1} & 10 & 100 &  1 & 10 & 100  & 1 & 10 & 100  & 1 & 10 & 100  & 1 & 10 & 100\\
  \midrule
  Equivariant Transporter                & \textbf{100} & \textbf{100} & \textbf{100} & \textbf{95.6} & \textbf{100} & \textbf{100} & \textbf{96.1} & \textbf{100} & \textbf{100} & \textbf{64.0} & \textbf{99.0} & \textbf{100} & \textbf{62.1} & \textbf{85.6} & \textbf{98.3} \\
  
  Transporter Network                  & 98.0 & \textbf{100} & \textbf{100}  & 82.3 & 94.8 & \textbf{100} &  84.2 & 99.6 & \textbf{100} & 45.0 & 85.0 & 99.0 & 16.6 & 63.3 & 75.0  \\
  
  \bottomrule
  \end{tabular}
  \vspace{0.5em}
  \caption{\scriptsize\textbf{Performance comparisons on tasks with a parallel-jaw end effector.} Success rate (mean\%) vs. the number of demonstration episodes (1, 10, or 100) used in training.}
  \vspace{-1.0em}
  \label{table:sample-efficiency2}
\end{table*}

\begin{table*}[ht]
  \setlength\tabcolsep{4pt}
  \centering
  \scriptsize
  \begin{tabular}{@{}l*{12}{>{\centering\arraybackslash}p{11mm}@{}}}
  \toprule
  & \multicolumn{3}{c}{goal-conditioned-block-insertion} & \multicolumn{3}{c}{goal-conditioned-block-pyramid} & \multicolumn{3}{c}{goal-conditioned-four-blocks} & \multicolumn{3}{c}{goal-conditioned-cable-align} \\
  \cmidrule(lr){2-4} \cmidrule(lr){5-7} \cmidrule(lr){8-10} \cmidrule(lr){11-13}
  Method & 1 & 10 & 100 &  1 & 10 & 100  & 1 & 10 & 100  & 1 & 10 & 100 \\
  \midrule
  Equivariant Transporter goal stack & \textbf{100} & \textbf{100} & \textbf{100} & \textbf{51.3} & \textbf{84.7} & \textbf{86.5} & \textbf{63.5} & \textbf{87.3} & \textbf{93.6} & \textbf{74.9} & \textbf{89.4} & \textbf{92.6}         \\
  

  Transporter goal stack & \textbf{100} & \textbf{100} & \textbf{100} & 9.8 & 64.7 & 72.5 & 11.2 & 24.1 & 25.3 &  49.5 &  78.7 &  88.7    \\
  
  Transporter goal split  & 99.0 & \textbf{100} & \textbf{100} & 3.3 & 58.8 & 67.7 & 17.8 & 27.3 & 27.9 & 52.5 &  84.4 & 92.1             \\
  
  \bottomrule
  \end{tabular}
  \vspace{0.5em}
  \caption{\scriptsize\textbf{Performance comparisons on goal-conditioned tasks.} Success rate (mean\%) vs. the number of demonstration episodes (1, 10, or 100) used in training.}
  \vspace{-1.0em}
  \label{table:goal-conditioned task}
\end{table*}
\subsection{Training and Evaluation}

\subsubsection{Training}

For each task, we produce a dataset of $n$ expert demonstrations, where each demonstration contains a sequence of one or more observation-action pairs $(o_t,\Bar{a}_t)$ (or $(o_t, o_g, \Bar{a}_t)$ for goal-conditioned tasks). Each action $\Bar{a}_t$ contains an expert picking action $\Bar{a}_{\pick}$ and an expert placing action $\Bar{a}_{\place}$. We use $\Bar{a}_t$ to generate one-hot pixel maps as the ground-truth labels for our picking model and placing model. The models are trained using a cross-entropy loss.

\subsubsection{Metrics}

We measure performance the same way as it was measured in Transporter Net~\cite{zeng2021transporter} -- using a metric in the range of 0 (failure) to 100 (success). Partial scores are assigned to multiple-action tasks. For example, in the block-stacking task where the agent needs to construct a 6-block pyramid, each successful rearrangement is credited with a score of 16.667. We report the highest validation performance during training, averaging over 100 unseen tests for each task. 

\subsubsection{Baselines} 

We compare our method against Transporter Net~\cite{zeng2021transporter} as well as the following baselines previously used in the Transporter Net paper~\cite{zeng2021transporter}. \textit{Form2Fit}~\cite{zakka2020form2fit} introduces a matching module with the measurement of $L_2$ distance of high-dimension descriptors of picking and placing locations. \textit{Conv-MLP} is a common end-to-end model~\cite{levine2016end} which outputs $a_{\pick}$ and  $a_{\place}$ using convolution layers and MLPs (multi-layer perceptrons). \textit{GT-State MLP} is a regression model composed of an MLP that accepts the ground-truth poses and 3D bounding boxes of objects in the environment. \textit{GT-State MLP 2-step} outputs the actions sequentially with two MLP networks and feeds $a_{\pick}$ to the second step. All regression baselines learn mixture densities~\cite{bishop1994mixture} with log-likelihood loss. 

For goal-conditioned tasks, we compare two baselines \textit{Equivariant-Transporter-Goal-Stack} with \textit{Transporter-Goal-Stack} and \textit{Transporter-Goal-Split}.

\begin{table*}[ht!]
  \setlength\tabcolsep{3pt}
  \centering
  \scriptsize
  \begin{tabular}{@{}l*{18}{>{\centering\arraybackslash}p{7mm}@{}}}
  \toprule
  Task Name-Demo Number & \multicolumn{3}{c}{Block Insertion-1} & \multicolumn{3}{c}{Block Insertion-10} & \multicolumn{3}{c}{Packing Boxes-1} & \multicolumn{3}{c}{Packing Boxes-10} & \multicolumn{3}{c}{Manipulating Rope-1} & \multicolumn{3}{c}{Manipulating Rope-10} \\
  \cmidrule(lr){2-4} \cmidrule(lr){5-7} \cmidrule(lr){8-10}\cmidrule(lr){11-13} \cmidrule(lr){14-16} \cmidrule(lr){17-19}
  
  Group Size & {$C_4$} & $C_6$ & $C_8$ & {$C_4$} & $C_6$ & $C_8$ & {$C_4$} & $C_6$ & $C_8$ & {$C_4$} & $C_6$ & $C_8$ & {$C_4$} & $C_6$ & $C_8$ & {$C_4$} & $C_6$ & $C_8$\\
  \midrule
  
  
  Best Mean Success Rate  &100 & 100 & 100 & 100 & 100 & 100 & 98.3 & 99.6  & 98.2 & 99.4 & 99.8 & 99.4 & 47.0 & 50.7 & 40.2 & 81.0 & 83.9 & 79.9 \\
  SGD Step of the Best Mean Success Rate & 4k & 1k & 1K & 1k & 1k &1k & 6k & 5k  & 6k & 2k & 8k & 6k & 6k & 7k & 7k & 1k & 4k &5k\\
  Mean Success Rate Trained with 1k Step & 92.0 & 100 & 100 & 100 & 100 & 100 & 87.7 & 94.1 & 97.2 & 99.2 & 97.2 & 98.8 & 28.0 & 39.0 & 12.3 & 81.0 & 70.9 & 43.2\\
  \bottomrule
  \end{tabular}
  \vspace{0.5em}
  \caption{\scriptsize\textbf{{Ablation Study on Group Size of Intermediate Equivariant Layers.}} We report the success rate (\% mean), and its corresponding training steps on three tasks from Ravens-10 Benchmark with 1 and 10 demos respectively. {We test three different cyclic groups ($C_4,C_6,C_8$) of the intermediate layers of the network $f_p,\psi$, and $\phi$.}}
  \vspace{-1.0em}
  \label{table:group_size_selection}
\end{table*}

\subsubsection{Adaptation of Transporter Net for Picking Using a Parallel Jaw Gripper}

In order to compare our method against Transporter Net for the five parallel jaw gripper tasks, we must modify Transporter to handle the gripper. {We accomplish this by~\citet*{zeng2018learning} lifting the input scene image $o_t$ over $C_n$, producing a stack of differently oriented input images which is provided as input to the pick network $f_{\pick}$. The results are then counter-rotated at the output of $f_{\pick}$ and each channel corresponds to one pick orientation.}

\subsection{Results for the Ravens-10 Benchmark Tasks}

\subsubsection{Task Success Rates} 

\begin{figure}
     \centering
     \begin{subfigure}[b]{0.45\textwidth}
         \centering
         \includegraphics[width=0.7\textwidth]{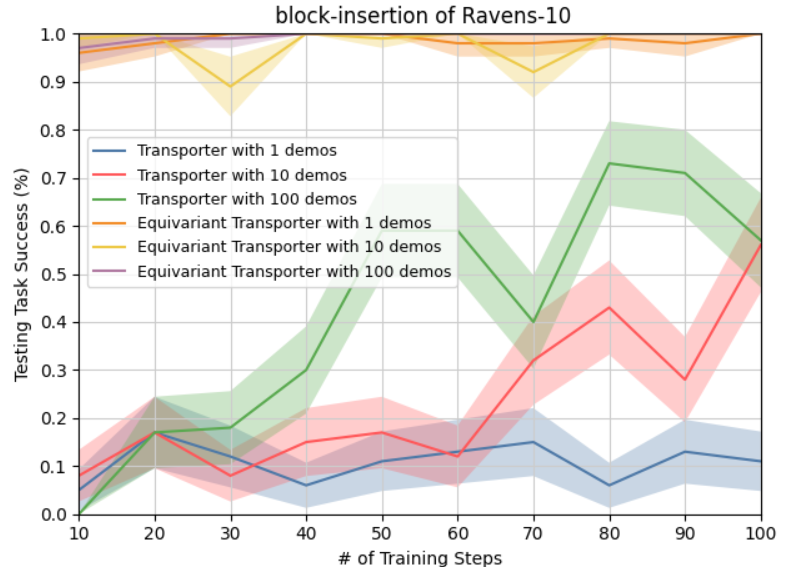}
     \end{subfigure}
     \hspace{0.5cm}
     \begin{subfigure}[b]{0.45\textwidth}
         \centering
         \includegraphics[width=0.7\textwidth]{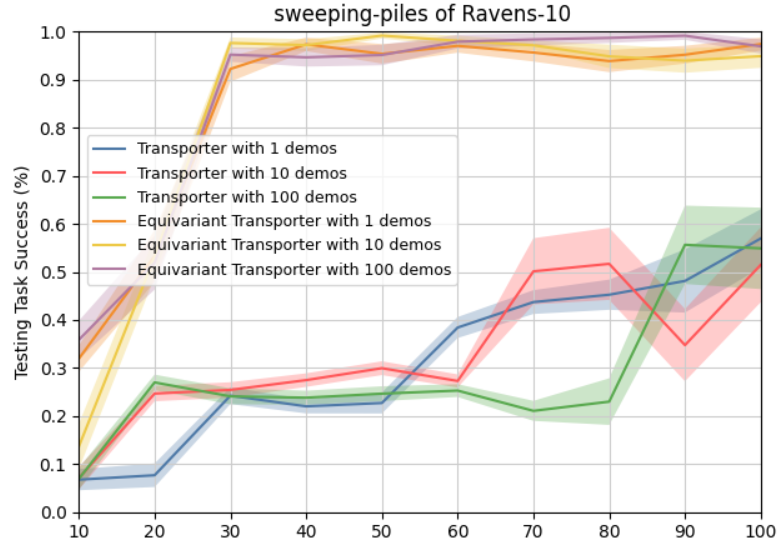}
     \end{subfigure}
     \caption[faster learning]{Equivariant Transporter Network converges faster than Transporter Network. Top: Block-insertion task. Bottom: sweeping-piles task. On the block insertion task, Equivariant Transporter can hit greater than $90\%$ success rate after 10 training steps and achieve $100\%$ success rate with less than 100 training steps.}
     \label{fig:fast_converge}
    \vspace{-0.5cm}
\end{figure}

Table~\ref{table:sample-efficiency1} shows the performance of our model on the Raven-10 tasks for different numbers of demonstrations used during training. All tests are evaluated on unseen configurations, i.e., random poses of objects, and three tasks (align-box-corner, assembling-kits, packing-box) use unseen objects. Our proposed Equivariant Transporter Net outperforms all the other baselines in nearly all cases. The amount by which our method outperforms others is largest when the number of demonstrations is smallest, i.e. with only 1 or 10 demonstrations. With just 10 demonstrations per task, our method can achieve $\geq 95\%$ success rate on 7/10 tasks. 
With either 1 or 10 demonstrations, the performance of our model is better than baselines trained with 1000 demonstrations on 5/10 tasks.
\subsubsection{Training Efficiency}

Another interesting consequence of our more structured model is that training is much faster. Figure~\ref{fig:fast_converge} shows task success rates as a function of the number of SGD steps for two tasks (Block Insertion and Sweeping Piles). Our equivariant model converges much faster in both cases. {It indicates that the large symmetry group enables the model to learn on a low-dimension space and achieve better convergence speed.}

\vspace{-0.1cm}
\subsection{Results for Parallel Jaw Gripper Tasks}

Table~\ref{table:sample-efficiency2} compares the success rate of Equivariant Transporter with the baseline Transporter Net on parallel-jaw gripper tasks. Again, our method outperforms the baseline in nearly all cases. One interesting observation that can be made by comparing Tables~\ref{table:sample-efficiency1} and~\ref{table:sample-efficiency2} is that both Equivariant Transporter and baseline Transporter do better on the gripper versions of the task compared to the original Ravens-10 versions. This is likely caused by the fact that the expert demonstrations we developed for the gripper version task have fewer stochastic gripper poses during pick than the case in the original Ravens-10 benchmark.

\subsection{Results for goal-conditioned equivariant transporter net}
Table~\ref{table:goal-conditioned task} compares the performance of \textit{Equivariant-Transporter-Goal-Stack} with the two baselines (\textit{Transporter-Goal-Stack, Transporter-Goal-Split}) for goal-conditioned tasks. Our model gets better performance than the baselines on all the tasks. In most cases, the performance gap between our model and the baselines becomes larger as the number of demonstrations decreases. It shows the sample efficiency of our proposed method could be used to solve goal-conditioned tasks effectively.

\subsection{Ablation Study}
\label{ablation_study}

\subsubsection{Ablations} 

We performed an ablation study to evaluate the relative importance of the equivariant models in pick ($f_p$ and $f_\theta$) and place ($\psi$ and $\phi$). We compare four versions of the model with various levels of equivariance: non-equivariant pick and non-equivariant place (baseline Transporter), equivariant pick and non-equivariant place, 
non-equivariant pick and equivariant place, and
equivariant pick and equivariant place (Equivariant Transporter). 

\begin{figure}[t]
    \centering
    \begin{subfigure}[b]{0.45\textwidth}
        \centering
        \includegraphics[width=0.7\textwidth]{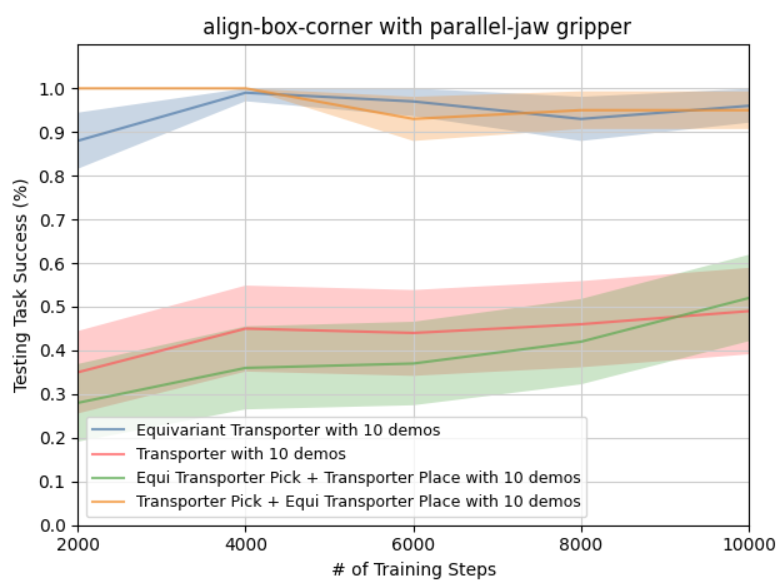}
    \end{subfigure}
    \vspace{0.1cm}
    \begin{subfigure}[b]{0.45\textwidth}
        \centering
        \includegraphics[width=0.7\textwidth]{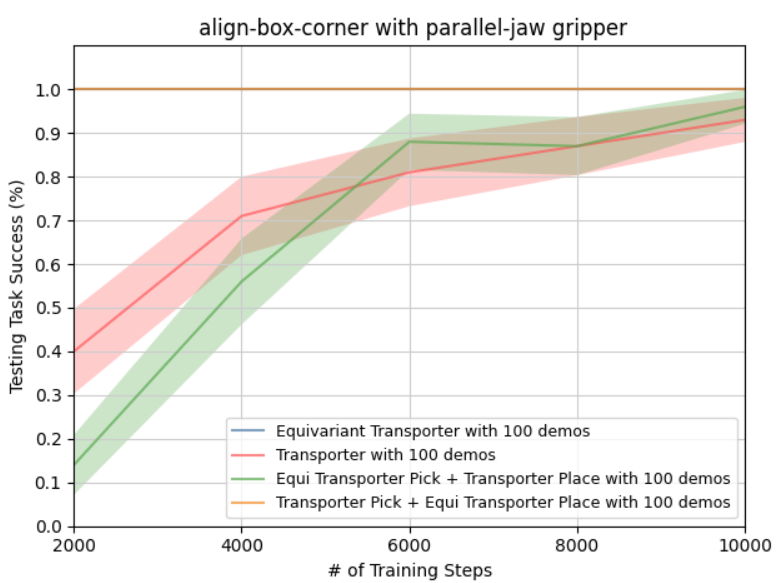}
    \end{subfigure}
    \caption[Ablation Study]{Ablation study. Performance is evaluated on 100 unseen tests of each task.}
    \label{fig:ablation}
\end{figure}

\subsubsection{Results} 

\figref{fig:ablation} shows the performance of all four ablations as a function of the number of SGD steps for the scenario where the agent is given 10 or 100 expert demonstrations. The results indicate that place equivariance (i.e. equivariance of $\psi$ and $\phi$) is namely responsible for the gains in performance of Equivariant Transporter versus baseline Transporter. This finding is consistent with the argument that the larger $C_n \times C_n$ symmetry group (only present with the equivariant place) is responsible for our performance gains. Though the non-equivariant and equivariant pick networks result in comparable performance, the equivariant network is far more computationally efficient. {The equivariant model takes a single image of the observation as input while the non-equivariant method~\citet*{zeng2018learning} needs a stack of n-different rotated input images in order to infer the pick orientation channel-wisely.}

\subsubsection{{Albation Study on Group Size}}
{We compare different group sizes to encode the latent features of our network. Note that the number of pick orientations and the number of place orientations are task-relevant parameters and this ablation study is used to investigate the group size of the intermediate layers. We select ($C_4,C_4,C_4$), ($C_6,C_6,C_6$), ($C_8, C_8,C_8$) to construct three different settings of $f_p,\psi$ and $\phi$. We build a light version of our model with each setting and train it on block insertion, packing boxes, and manipulating rope with 1 and 10 demos. Specifically, the $C_4$ model has 9 million trainable parameters, and the $C_6$ model and the $C_8$ model have 13.5 million and 18 million parameters, respectively. Note that a large group has a large fiber space dimension which results in more parameters but it also adds more constraints to the free parameters of the kernel. We train each model with data augmentation and evaluate the performance on 100 unseen test cases every 1k training steps. We report the highest success rate, its corresponding training steps, and the success rate with 1k training steps in Table~\ref{table:group_size_selection}. Several findings could be drawn from Table~\ref{table:group_size_selection}. First, the best mean success rates are consistent with different groups on most tasks. As the number of available demonstrations increases, the differences decrease. Second, large group size may boost the convergence speed when looking at the results of Block Insertion-1. However, it could also result in overfitting when comparing the results of Manipulating-Rope-1 since the large-group model has more constraints and parameters. Finally, we think the group size of the intermediate layer might be regarded as a hyper-parameter.}

\subsection{Experiments on a Physical Robot}

We evaluated Equivariant Transporter on a physical robot in our lab. The simulator was not used in this experiment -- all demonstrations were performed on the real robot.

\subsubsection{Setup} 

\begin{figure}[t]
     \centering
     \begin{subfigure}[b]{0.23\textwidth}
         \centering
         \includegraphics[width=\textwidth]{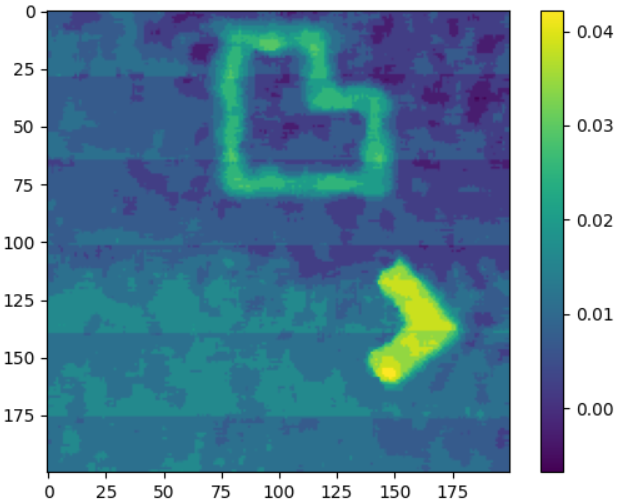}
     \end{subfigure}
     \hspace{0.1cm}
     \begin{subfigure}[b]{0.23\textwidth}
         \centering
         \includegraphics[width=\textwidth]{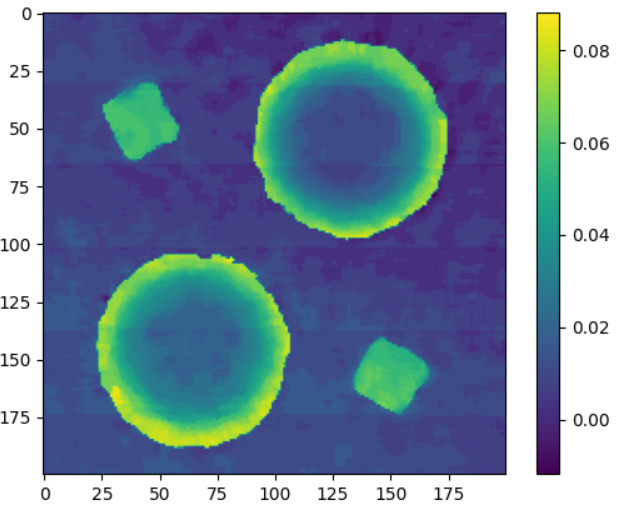}
     \end{subfigure}
     \caption[depth image]{Real robot experiment: initial observation $o_t\in\mathbb{R}^{200\times200}$ from the depth sensor. The left figure shows the block insertion task; the right figure shows the task of placing boxes in bowls. The depth value ($\mathrm{meter}$) is illustrated in the color bar.}
     \label{fig:depth}
\end{figure}

We used a UR5 robot with a Robotiq-85 end effector. The workspace was a $40cm \times 40cm$ region on a table beneath the robot (see Figure~\ref{fig:robot}). The observations $o$ were $200 \times 200$ depth images obtained using an Occipital Structure Sensor that was mounted pointing directly down at the table (see Figure~\ref{fig:depth}). 

\subsubsection{Tasks}

We evaluated Equivariant Transporter on three of the Ravens-10 gripper-modified tasks: block insertion, placing boxes in bowls, and stacking a pyramid of blocks. Since our sensor only measures depth (and not RGB), we modified the box-in-bowls task such that box color was irrelevant to success, i.e. the task is simply to put any box into a bowl.

\subsubsection{Demonstrations}

We obtained 10 human demonstrations of each task. These demonstrations were obtained by releasing the UR5 brakes and pushing the arm physically so that the harmonic actuators were back-driven.

\begin{figure}[t]
    \centering
    \includegraphics[width=0.45\textwidth]{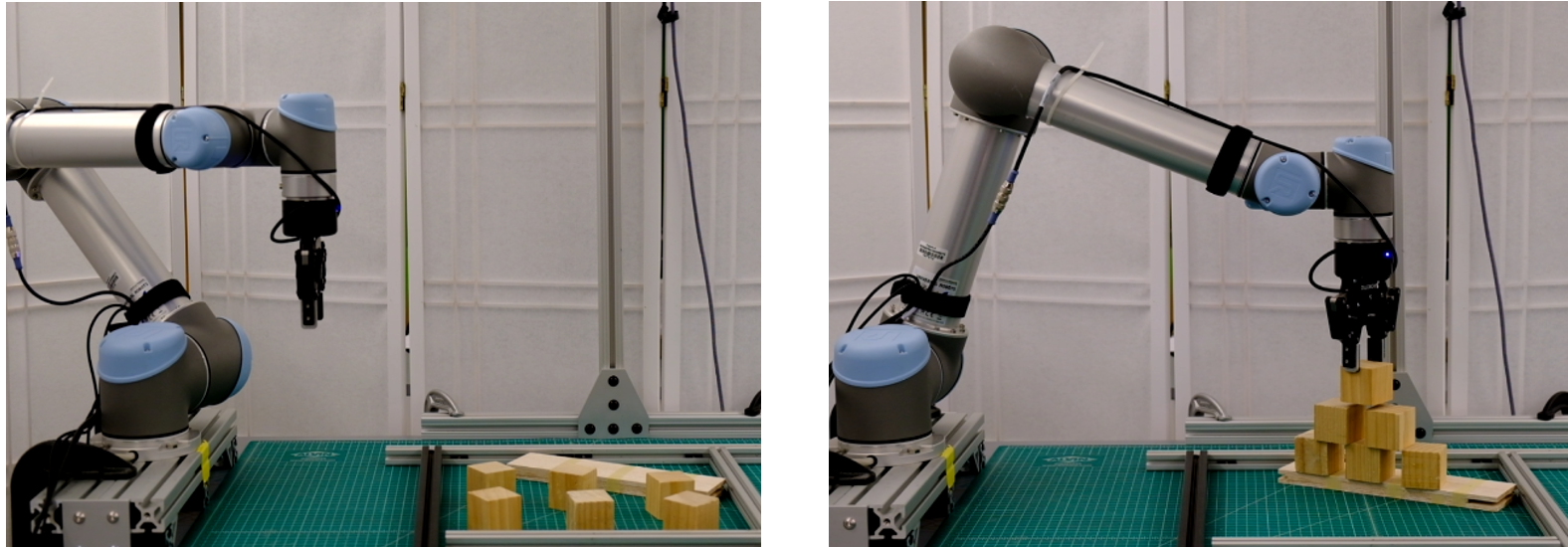}
    \caption{Stack-block-pyramid task on the real robot. The left figure shows the initial state; the right figure shows the completion state.}
    \label{fig:robot}
    \vspace{-0.3cm}
\end{figure}
\begin{table}[ht]
    \centering
    \scriptsize
    \begin{tabular}{c|c|c|c}
    \hline
         Task & \# demos & \# completions / \# trials  & success rate\\
        \hline
         stack-block-pyramid & 10 & 17/20  & 95.8\%\\
        \hline
        place-box-in-bowl & 10 & 20/20  & 100\%\\
        \hline
        block-insertion & 10 & 20/20  & 100\%\\
     \hline
    \end{tabular}
    \setlength{\belowcaptionskip}{0.2cm}
    \caption{Task success rates for physical robot evaluation tasks.}
    \label{tab:real_test_performance}
\end{table}

\subsubsection{Training and Evaluation}

For each task, {a single-policy agent} was trained for 10k SGD steps. During testing, objects were randomly placed on the table. A task was considered to have failed when a single incorrect pick or place occurred. We evaluated the model on 20 unseen configurations of each task.

\subsubsection{Results}
Table\:\ref{tab:real_test_performance} shows results from 20 runs of each of the three tasks. Notice that the success rates here are higher than they were for the corresponding tasks performed in simulation (Table~\ref{table:sample-efficiency2}). This is likely caused by the fact that the criteria for task success in simulation (less than 1 cm translational error and less than $\frac{\pi}{12}$ rotation error were more conservative than is actually the case in the real world.

\subsection{{Discussion}}
{Equivariant networks are built on top of conventional convolution kernels with the steerability constraint. It does not break the mechanism of weight sharing and updating and thus keeps the robustness of learning and reasoning of CNN. As shown in Figure~\ref{fig:depth}, Equivariant Transporter Net can handle real-sensor noise. Frequently, the crop $c$ contains multiple objects. For instance, on the stack-block-pyramid task as shown in Figure~\ref{fig:robot}, the crop not only includes the block to be picked but also neighboring blocks or some parts of them. During training, data augmentation also generates images with partially observed objects. For example, on the block insertion task, it shifts some part of the L-shape block or the slot out of the scene.
Some special shapes like elongated objects might be difficult to represent with an image crop and may be suitable for the goal-conditioned version of our model. Some high-precision tasks like gear assembly are more sensitive to discretization and it may be tackled more easily with the $\SOtwo$ version of our model.}

{
Compared to learning pick-and-place skills efficiently, the one-shot generalization and sequential decision-making ability of both Transporter Net and Equivariant Transporter Net seem less compelling. As shown in Table~\ref{table:sample-efficiency1}, they achieved less than $50\%$ success rate when trained with 1 demo on the align-box-corner task that requires the agent to generalize the skill to boxes with random sizes and colors during the test. The performances on the stack-block-pyramid task trained with 1 demo are below $40\%$ since if there was a collapse, some blocks might be tilted and it yields out-of-distribution data.
}

\section{Conclusion and Limitations}
\label{sec:conclusion}

This paper explores the symmetries present in the pick and place problem and finds that they can be described by pick symmetry and place symmetry. This corresponds to the group of different pick and place poses. We evaluate the Transporter Network model proposed in~\citet*{zeng2021transporter} and find that it encodes half of the place symmetry (the $C_n$-place symmetry). We propose a novel version of Transporter Net, Equivariant Transporter Net, which we show encodes both types of symmetries. The large symmetry group could also be extended to solve goal-conditioned tasks. We evaluate our model on the Ravens-10 Benchmark and its variations and evaluate against multiple strong baselines. Finally, we demonstrate that the method can effectively be used to learn manipulation policies on a physical robot. 

One limitation of our framework as it is presented in this paper is that it relies entirely on behavior cloning. A clear direction is to integrate more on-policy learning which we believe would enable us to handle more complex tasks. {Other directions of the multi-task language-conditioned equivariant agent, a closed-loop policy, or 3D Equivariant Transporter Net are also interesting.}

\bibliography{reference}
\end{document}